\documentclass[10pt,journal,compsoc]{IEEEtran}

\ifCLASSOPTIONcompsoc
  \usepackage[nocompress]{cite}
\else
  \usepackage{cite}
\fi

\usepackage{amsmath}
\usepackage{algorithm}
\usepackage{bbm}
\usepackage{dsfont}
\usepackage{graphicx}  
\usepackage{epstopdf}
\usepackage{amsfonts}
\usepackage{algpseudocode}
\usepackage{multirow}
\usepackage{subfigure}
\usepackage{hyperref}
\usepackage{amssymb}

\usepackage{longtable}
\usepackage[small]{caption}
\ifCLASSINFOpdf

\else

\fi

\begin{document}

\title{Multi-label Classification with High-rank and High-order Label Correlations }

\author{Chongjie Si,
        Yuheng Jia,~\IEEEmembership{~Member,~IEEE},
        Ran Wang,~\IEEEmembership{Senior~Member,~IEEE},
        Min-Ling Zhang,~\IEEEmembership{Senior~Member,~IEEE},
        Yanghe Feng,
        Chongxiao Qu

\IEEEcompsocitemizethanks{
\IEEEcompsocthanksitem This work was supported in part by the National Natural Science Foundation of China under Grant 62106044, 62176160 and 62225602, in part by the Natural Science Foundation of Jiangsu Province under Grant BK20210221, in part by the Guangdong Basic and Applied Basic Research Foundation (Grant 2022A1515010791), in part by ZhiShan Youth Scholar Program from Southeast University under Grant 2242022R40015, and in part by the Hong Kong UGC under grant UGC/FDS11/E02/22. (Corresponding author: Yuheng Jia.)

\IEEEcompsocthanksitem C. Si is with the Chien-Shiung Wu College, Southeast University, Nanjing 210096, China, and also with the MoE Key Lab of Artificial Intelligence, AI Institute, Shanghai Jiao Tong University, Shanghai 200240, China. \protect E-mail: chongjiesi@sjtu.edu.cn.

\IEEEcompsocthanksitem Y. Jia is with the School of Computer Science and Engineering, Southeast University, Nanjing 210096, and with the Key Laboratory of New Generation Artificial Intelligence Technology and Its Interdisciplinary Applications (Southeast University), Ministry of Education, China, and also with School of Computing \& Information Sciences, Caritas Institute of Higher Education, Hong Kong. \protect
E-mail: yhjia@seu.edu.cn.

\IEEEcompsocthanksitem M. Zhang is with the School of
Computer Science and Engineering, Southeast University, Nanjing 210096, China, and also with the Key Laboratory of Computer Network and Information Integration (Southeast University), Ministry of Education, China \protect
e-mail: zhangml@seu.edu.cn.

\IEEEcompsocthanksitem R. Wang is with the School of Mathematical Science, Shenzhen University, Shenzhen 518060, China, and also with Shenzhen Key Laboratory of Advanced Machine Learning and Applications, Shenzhen University, Shenzhen 518060, China. \protect E-mail: wangran@szu.edu.cn.

\IEEEcompsocthanksitem Y. Feng is with the College of Systems Engineering, National University of Defense Technology.\protect E-mail: fengyanghe@nudt.edu.cn.

\IEEEcompsocthanksitem C. Qu is with the 52nd Research Institute of China Electronics Technology Group.\protect E-mail:quchongxiao@163.com.} 
}

\markboth{Journal of \LaTeX\ Class Files,~Vol.~14, No.~8, August~2015}%
{Shell \MakeLowercase{\textit{et al.}}: Bare Demo of IEEEtran.cls for Computer Society Journals}

\IEEEtitleabstractindextext{%
\begin{abstract}
Exploiting label correlations is important to multi-label classification. Previous methods capture the high-order label correlations mainly by transforming the label matrix to a latent label space with low-rank matrix factorization. However, the label matrix is generally a full-rank or approximate full-rank matrix, making the low-rank factorization inappropriate. Besides, in the latent space, the label correlations will become implicit. To this end, we propose a simple yet effective method to depict the high-order label correlations explicitly, and at the same time maintain the high-rank of the label matrix. Moreover, we estimate the label correlations and infer model parameters simultaneously via the local geometric structure of the input to achieve mutual enhancement. Comparative studies over twelve benchmark data sets validate the effectiveness of the proposed algorithm in multi-label classification. The exploited high-order label correlations are consistent with common sense empirically. \textbf{Our code is publicly available at \url{https://github.com/Chongjie-Si/HOMI}.}
\end{abstract}

}

\maketitle

\IEEEdisplaynontitleabstractindextext

\IEEEpeerreviewmaketitle

\IEEEraisesectionheading{\section{Introduction}\label{sec:introduction}}

\IEEEPARstart{R}{ecently}, multi-label classification has attracted a lot of attention, aiming to solve real-world tasks with rich semantics  \cite{rubin2012statistical}, \cite{sun2014multi}, \cite{wang2013multiqghjerhw}, \cite{WANG2021107583}. Specifically, in multi-label classification, one instance may be associated with several labels. For example, an image may be associated with a set of tags \cite{BOUTELL2004Learning}, and a piece of news may belong to several topics. Different from the traditional single-label classification problem which can be regarded as a degenerated version of multi-label classification, the overwhelming size of output space makes multi-label classification a much more challenging task.

Exploiting label correlations is of great importance, as label correlations can provide valuable semantic relationship for the output of multi-label classification. For instance, if two labels, ``rainforest" and ``soccer" are assigned to a sample, then the label ``Brazil" may be also assigned to it. Similarly, if ``teacher" and ``blackboard" are present, it is very likely that label ``classroom" will also be present. Based on how to explore the label correlations,  multi-label classification methods can be roughly divided into three families: first-order, second-order and high-order. For first-order methods, the label correlations are not considered. For example, binary relevance (BR)  \cite{BOUTELL2004Learning} transformed multi-label classification into a set of independent binary classification problems and solved them separately. The second-order methods take the pairwise relationship of labels into considerations. For example, multi-label classification with Label specIfic FeaTures (LIFT) \cite{2011Lift} employed clustering techniques to find second-order correlations between labels. However, in real-world scenarios, the label correlations may be much more complex than first-order and second-order relations. To this end, many high-order label correlations exploiting methods were proposed. For example, classifier chains (CC) \cite{2009Classifier1212} transformed the multi-label classification into a chain of binary classification problems. Random-k-labelsets (RAKEL) \cite{tsoumakas2010random} converted multi-label classification into a set of multi-class classification problems over k randomly-chosen class labels. Some approaches assumed the labels were correlated in a hierarchical structure \cite{punera2005automaticallyqgw}. All the mentioned approaches specify the high-order correlations of labels manually, which will depress the classification performance if the manual setting is inappropriate.

Recently, some high-order approaches were proposed to automatically explore the high-order label correlations \cite{2017Multisdfsaf}, \cite{2019collaboration}. They generally decomposed the label matrix to a latent space by low-rank matrix factorization \cite{9336710}, and then assumed the latent labels may capture the higher level semantic concepts \cite{8869493}. \textit{However, as can be seen from Table \ref{tab:rankandnumber}, the rank of the label matrix usually equals to or approximately equals to the number of labels, which means the label matrix is full-rank or approximate full-rank, making the low-rank matrix assumption inappropriate. Besides, in the latent space, the label correlations become indirect and semantically unclear. 
}
\begin{table*}[ht]\scriptsize
\setlength{\tabcolsep}{1mm}\renewcommand\arraystretch{1} 
    \centering
 \caption{The rank of the label matrix on some commonly used multi-label data sets.}
    \begin{tabular}{cc c c c c c c c c c c c}
     \hline \hline
        Data set & mediamill & CAL500 & emotions & enron & bibtex & delicious & language log & birds & yeast & scene & corel5k & corel16k \\
        Number of labels &101 & 174 & 6 & 53 & 159 & 983 & 75 & 19 & 14 & 6 & 374 & 153\\
        Number of samples & 43097 & 502 & 593 & 1702 & 7395 & 16105 & 1460 & 645 & 2417 & 2407 & 5000 & 13766 \\ 
        Size of label matrix & 43907$\times$101 & 502$\times$174 & 593$\times$6 & 1702$\times$53 & 7395$\times$159 & 16105$\times$983 & 1460$\times$75 & 645$\times$19 &2417$\times$17 &2407$\times$6 & 5000$\times$374 &13766$\times$153\\ 
        Rank of label matrix & 100 & 174 & 6 & 52 & 159 & 983 & 75 & 19 & 14 & 6 & 371 & 153 \\
        
        \hline\hline
    \end{tabular}
   
    \label{tab:rankandnumber}
\end{table*}

To solve these issues, in this paper, we propose a novel approach called HOMI, with High-Rank and \textbf{H}igh-\textbf{O}rder \textbf{M}ult\textbf{I}-Label learning. Specifically, we argue that if a label is highly correlated to a set of other labels, it can be easily reconstructed by the linear combination of that set of labels. Therefore, we propose to use self-representation to exploit the high-order label correlations for multi-label classification. Note that it can keep the rank of the label matrix unchanged and indicate the high-order correlations among labels explicitly. Moreover, the local geometric structure is also beneficial to multi-label classification, as if two samples are similar to each other in the feature space, they are likely to share similar labels. Here, we adopt an $s$-nearest-neighbors (SNN) graph to depict the local geometric structure of the input samples and incorporate the local geometric structure by a graph Laplacian regularization. Besides, the proposed approach naturally unifies high-order label correlations learning and multi-label prediction into a joint model via the graph Laplacian regularization, such that those two separate processes can be well interacted with each other to achieve mutual enhancement. Comprehensive experiments substantiate that HOMI outperforms the state-of-the-art multi-label classification methods significantly, and reasonable high-order label correlations can be constructed by HOMI. 

The rest of the paper is organized as follows. We first review some related works in exploiting label correlations and explain why the label matrix should be full-rank or approximate full-rank in multi-label classification in Section 2. Then, we introduce the proposed approach as well as the numerical solution in Section 3, and present the experimental results and analysis in Section 4. Finally, conclusion is given in Section 5.

\section{Related Work}
\subsection{Exploiting Label Correlations in Multi-label Classification }
In multi-label classification, an instance is associated with a set of labels. In recent years, this new machine learning paradigm has made great progress and has been widely used in image classification \cite{0Multi}, \cite{2015Multi}, \cite{2008Automatic}, automatic annotation \cite{2009Multi}, \cite{2010Multi}, \cite{2010Transductive}, web mining \cite{2012Unsupervised}, \cite{2009Towards}, \cite{2015Classification}, audio recognition \cite{wu2014music}, \cite{2011Cost}, \cite{2009Improving}, \cite{2006Multi} and information retrieval \cite{2012Multilabel}, \cite{2016Ensemble}, etc. 

However, the task of inducing multi-label classification functions is challenging, as the classifier's output space is exponential in size to the number of possible class labels, i.e., $|2^\mathcal{Y}|$, where $\mathcal{Y}$ denotes the number of possible labels. A useful way to cope with this issue is to exploit label correlations to simplify the learning procedure. Based on the degree of label correlations used, the algorithms of multi-label classification can be divided into three categories \cite{2010Multilabeldependency}: first-order, second-order and high-order.
 
First-order methods do not take label correlations into consideration and assume that all the labels are independent. BR \cite{BOUTELL2004Learning} is a prevailing first-order approach, transforming the original multi-label classification into a set of independent binary classification tasks. ML-KNN \cite{zhang2007mlds} is also a popular first-order algorithm based on k-nearest-neighbour classification. The major advantage of first-order approaches is the conceptual simplicity and high efficiency, for they are easy to understand and operate. Nevertheless, they ignore the label correlations, which results in performance degeneration.
 
Second-order approaches focus on pairwise label relations. For instance, calibrated label ranking (CLR) \cite{Johannes2008Multilabel} and LIFT \cite{2011Lift} are two representative approaches, transforming original multi-label classification into pairwise ranking problems. Second-order approaches are relatively more effective than first-order ones in exploiting label correlations. However, in real-world applications, the relationship of labels may be quite complex and sophisticated, such that the pairwise relationship cannot describe the real-world label correlations very well.
 
High-order approaches aim to dig the high-order label correlations. CC \cite{2009Classifier1212}, for instance, converted the multi-label task into a chain of independent binary classification problems, with the ground-truth labels decoded into the features each time. RAKEL \cite{tsoumakas2010random} reformulated the multi-label classification into several sets of multi-class classification tasks.

Recently, some low-rank based approaches proposed to learn high-order label correlations based on the assumption that the label matrix is low-rank, for there are correlations among labels in multi-label classification. For example, Zhu et al. \cite{2017Multisdfsaf} used low-rank decomposition in multi-label learning, exploiting global and local label correlations simultaneously, through learning a latent label representation and optimizing label manifolds \cite{9007480}. Wang et al. \cite{9084698} controlled the sparsity of the coefficient matrix to filter out label-specific features and applied low-rank constraints to the label matrix to mine the local correlations of class labels. Xu et al. \cite{xu2014learningssdsdsd} proposed an integrated framework that learns the correlations among labels while training the multi-label model simultaneously, and specifically adopted a low-rank structure to capture the complex correlations among labels. Moreover, Yu et al. \cite{yu2014largdffafae} proposed an approach learning a linear instance-to-label mapping with low-rank structure, and implicitly taking advantage of global label correlations.

The general strategy of the above mentioned methods to capture label semantics is to decompose the label matrix to a latent label space by low-rank matrix factorization, \cite{2017Costsfds}, \cite{2017Learningsfsfa}. Specifically, denote the label matrix $\mathbf{Y} \in \mathbb{R}^{ n\times l}$, with $n$ and $l$ being the number of samples and labels, respectively. The low-rank based approaches usually decompose $\mathbf{Y}$ into two smaller matrices $\mathbf{U}$ and $\mathbf{V}$, i.e., $\mathbf{Y}$ can be written as the product

\begin{equation}
    \min_\mathbf{UV}\|\mathbf{Y} - \mathbf{UV} \|_F^2,
    \label{eq lowrank}
\end{equation}
where $\mathbf{V} \in \mathbb{R}^{h\times l}$ is the latent label matrix exploiting higher level label semantics, and $\mathbf{U}\in \mathbb{R}^{n\times h}$ is a basis matrix relating the original labels to the latent labels. As $h$ is smaller than $l$ and $n$, the rank of $\mathbf{UV}$ is smaller than $\mathbf{Y}$, i.e., Eq. (\ref{eq lowrank}) approximates the label matrix by low-rank factorization.
 
\subsection{Why is the label matrix full-rank?}

However, we believe that the label matrix is full-rank or approximately full-rank, and accordingly the low-rank matrix assumption is not the best choice for multi-label learning. The reasons are as follows. First, the size of the label matrix $\mathbf{Y}$ is $ n\times l$, as $l \ll n$ in general, the rank of $\mathbf{Y}$ is very likely to be or close to $l$. Second, although there are some connections among labels, the connections are usually not determinated. For example, if label A is related to label B, in other words, if A appears on a sample, we can infer that B has high possibility to be also appeared on that sample. But we cannot conclude that B will absolutely appear, as there will always be samples that only have label A or B. Therefore, those connections cannot reduce the rank of the label matrix, and likewise cannot result in a low-rank label matrix. Last but not least, as shown in the Table \ref{tab:rankandnumber}, the commonly used multi-label data sets are always full-rank or approximately full-rank, which further empirically validates that the label matrix of multi-label classification should be high-rank rather than low-rank. 

As a summary, the label matrix in multi-label learning is usually a full-rank matrix, which cannot be well approximated by low-rank decomposition. Besides, in the latent space, the label correlations become indirect and semantically unclear. To solve these issues, in the next section, a new approach named HOMI is proposed, which can keep the rank of the label matrix unchanged, and indicate label correlations directly in the label space.

\subsection{Deep-learning Based Multi-label Classification}
Due to its robust learning capability, deep learning has emerged as an important technique for achieving multi-label classification. In those methods, effectively leveraging deep learning is critical for capturing intricate label dependencies. To exploit the underlying intricate label structure, Cisse et al. (2016) proposed ADIOS \cite{cisse2016adios}, which employs a novel deep architecture that partitions the labels into a Markov blanket chain, capitalizing on this partition to enhance classification performance. Wang et al. (2016) introduced CNN-RNN \cite{wang2016cnn}, which leverages recurrent neural networks (RNNs) to better model higher-order label dependencies. CNN-RNN learns a unified image-label embedding that encapsulates both semantic label dependencies and image-label relevance. Notably, this approach enables end-to-end training from scratch. Moreover, Nam et al. proposed an alternative technique to the traditional classifier chain method \cite{nam2017maximizingseq2seq}. Their approach employs RNNs to convert the MLC problem into a sequential prediction task, with initially arbitrary label ordering. This method offers the advantage of focusing on predicting positive labels exclusively, significantly reducing the prediction space compared to the complete set of labels.

These methodologies exemplify diverse strategies for harnessing deep learning in MLC, each addressing label dependencies through distinctive means. However, different from these methods, HOMI explicitly reveals the label dependency based on the full-rank assumption, which further improves the MLC problem.

\section{The Proposed Approach}
Motivated by the fact that the label matrix is generally full-rank, in this section, we introduce HOMI to exploit the high-order label correlations for multi-label classification.  
Prior to that, we first briefly summarize the notations used in this paper. Formally, let $\mathcal{X} = \mathbb{R}^m$ denote the $m$-dimensional feature space and $\mathcal{Y} = \{c_1, c_2,..,c_l\}$ denote the label space of $l$ labels, where $c_i\in\{0,1\}$ stands for the $i$th label, multi-label classification learns a function ${f}: \mathcal{X}\rightarrow 2^{\mathcal{Y}}$ from the training data set $\mathcal{D} = \{( \mathbf{x}_i, \mathbf{y}_i)|1\le i\le n\}$, where $\mathbf{x}_i$ stands for the $i$th instance and $\mathbf{y}_i$ stands for the corresponding label set, and $n$ is the number of instances. Let $ \mathbf{X} = [\mathbf{x}_1,\mathbf{x}_2,...,\mathbf{x}_n]^\mathsf{T} \in \mathbb{R}^{n\times m} $ denote the instance matrix and $\mathbf{Y}= [\mathbf{y}_1,\mathbf{y}_2,...,\mathbf{y}_n]^\mathsf{T} \in \{0,1\}^{n\times l}$ denote the label matrix with $l$ labels. Note that the value of the labels is binary, and we have $\mathbf{y}_i = [y_{i1}, y_{i2},...,y_{il}]$ where $y_{ij} = 1$ (resp. $y_{ij}=0$) if the sample $\mathbf{x}_i$ has (resp. does not have) the $j$th label.

\subsection{Basic Model}

First, HOMI uses a weight matrix $\mathbf{W}=[\mathbf{w}_1,\mathbf{w}_2,...,\mathbf{w}_l]\in \mathbb{R}^{m \times l}$ to map the instance to the labels by minimizing the following least squares loss:

\begin{equation}
\begin{split}
    \min_{\mathbf{W},\mathbf{z}}
    \|\mathbf{Y}-\mathbf{XW}-\mathbf{1}_n\mathbf{z}^\mathsf{T}\|^2_F +\lambda(\|\mathbf{W}\|^2_F+\|\mathbf{z}\|^2_2),
    \end{split}
    \label{basic model}
\end{equation}
where $\|\cdot\|_F$ and $\|\cdot\|_2$ stand for the Frobenius norm and $\ell_2$ norm of a matrix and a vector, respectively. $\mathbf{z} = [z_1, z_2, ... , z_l]^\mathsf{T}\in \mathbb{R}^{l}$ is the bias term and $\mathbf{1}_n\mathbf{}\in \mathbb{R}^{n}$ is an all one vector. The first term in Eq. (\ref{basic model}) measures the predictive error of the model, and the second term is the regularization of the weight matrix $\mathbf{W}$ and the bias $\mathbf{z}$, trying to control the complexity of the whole model, and $\lambda \geq 0$ is a hyper-parameter to balance those two terms.

\subsection{Exploiting High-Order Label Information}
As mentioned earlier, if a label is highly correlated with other labels, it can be easily reconstructed by those labels. Thus, we propose to adopt the self-representation strategy to dig the high-order correlations between labels, which can be mathematically formulated as

\begin{equation}
    \begin{split}
        \min_{\mathbf{B, t}}
        \|\mathbf{Y}-\mathbf{YB}-\mathbf{1}_n\mathbf{ t}^\mathsf{T}\|_F^2 + \lambda(\|\mathbf{B}\|^2_F +  \|\mathbf{t}\|^2_2),
    \end{split}
    \label{high-order inf}
\end{equation}
where $\mathbf{B}\in \mathbb{R}^{l\times l}$ records the high-order label correlations, and similar to Eq. (\ref{basic model}), $\mathbf{t}\in \mathbb{R}^{l}$ is the bias term for better self-regression. We also introduce an additional penalty term on $\mathbf{B}$ and $\mathbf{t}$ to avoid the trivial solution (i.e., $\mathbf{B}=\mathbf{I}$ with $\mathbf{I}$ being an identity matrix) and over-fitting. 

The previous methods use low-rank matrix factorization to decompose the label matrix $\mathbf{Y}$ to a latent space to exploit the high-order correlations, however, the label matrix is usually a full-rank matrix. Technically, this is because although correlations exist in labels, the correlated labels also have a chance to exist alone on some samples.  Table \ref{tab:rankandnumber} also empirically verifies this observation. The full-rank property of the label matrix makes the low-rank factorization-based methods unreasonable. Differently, the adopted self-representation approach can keep the rank of label matrix unchanged \cite{elhamifar2016high}. 
Moreover, in the latent space, the label correlations can only be captured implicitly, while on the contrary, the elements in $\mathbf{B}$ can explicitly indicate the correlations between two labels.

\subsection{Incorporating Local Geometric Structure}

HOMI also takes the local geometric structure of instances into consideration, i.e., if $\mathbf{x}_i$ is similar to $\mathbf{x}_j$, the predictive label sets of them may have some labels in common. Specifically, we first calculate the Pearson correlation coefficient of samples, i.e., 
$$\mathbf{R}_{ij} = \frac{{\rm Cov}(\mathbf{x}_i,\mathbf{x}_j)}{\sigma_{\mathbf{x}_i}\sigma_{\mathbf{x}_j}},$$
where ${\rm Cov}(\mathbf{x}_i,\mathbf{x}_j)$ is the covariance of $\mathbf{x}_i$ and $\mathbf{x}_j$, and $\sigma_{\mathbf{x}_i}$ is the standard deviation of $\mathbf{x}_i$. Afterwards, we construct an $s$-nearest-neighbors graph to capture the local geometric structure of the input, i.e., 

\begin{equation}
    \mathbf{S}_{ij} = 
    \begin{cases}
     \mathbf{R}_{ij} ~& {\rm if}(i,j) \in \mathcal{N}_{si} \\
    0 ~& {\rm otherwise}
    \end{cases},
    \label{sij}
\end{equation}
where $\mathcal{N}_{si}$ is the set of $s$ nearest instances of the $i$th instance, and we choose the ones with the top $s$ values of $\mathbf{R}_{ij}$ of the $i$th sample as its neighbors. Then, the local geometric structure is incorporated by minimizing 
\begin{equation}
    \sum_{i,j} \mathbf{S}_{ij}\|\mathbf{g(x}_i\mathbf{)} - \mathbf{g(x}_j\mathbf{)}\|^2_2={\rm tr}(\mathbf{K}^\mathsf{T}\mathbf{LK}),
    \label{geometric}
\end{equation}
where $\mathbf{L}=\mathbf{Q}-\mathbf{S}$ is the graph Laplacian matrix with $\mathbf{Q}$ being the diagonal degree matrix of $\mathbf{S}$. In order to learn the local structural information more reasonably, we make $\mathbf{L}$ symmetric, i.e., $\mathbf{L}=\frac{1}{2}(\mathbf{L}+\mathbf{L}^\mathsf{T})$. $\mathbf{g}$ is the discriminative function, i.e. $\mathbf{g(x}_i\mathbf{)} = (\mathbf{x}_i\mathbf{W}+\mathbf{z}^\mathsf{T})\mathbf{B} + \mathbf{t}^\mathsf{T}$, which takes the high-order label correlations into account. $\mathbf{K} = [\mathbf{g(x}_1\mathbf{)}, \mathbf{g(x}_2\mathbf{)},...,\mathbf{g(x}_n\mathbf{)}]^\mathsf{T}\in \mathbb{R} ^{n\times l}$ (i.e. $\mathbf{K} = (\mathbf{XW} + \mathbf{1}_n\mathbf{z}^\mathsf{T})\mathbf{B} + \mathbf{1}_n\mathbf{t}^\mathsf{T}$). If Eq. (\ref{geometric}) is minimized, two similar instances will have similar predictive label sets. 

\subsection{Model Formulation}
Based on the above discussion, the objective function of HOMI is finally formulated as
\begin{equation}
    \begin{split}
        \min_{\mathbf{W,B,z,t}}  & \|\mathbf{Y}-\mathbf{XW}-\mathbf{{1_n}z}^\mathsf{T}\|^2_F + \gamma{\rm tr}(\mathbf{K}^\mathsf{T}\mathbf{LK})\\ &
        +\beta\|\mathbf{Y}-\mathbf{YB}-\mathbf{{1_n}t}^\mathsf{T}\|_F^2
        +\lambda(\|\mathbf{W}\|^2_F \\ &  +\|\mathbf{z}\|^2_2+\|\mathbf{B}\|^2_F+ \|\mathbf{t}\|^2_2),
    \end{split}
    \label{loss}
\end{equation}
where $\gamma$, $\beta$, and $\lambda$ denote different hyper-parameters to balance different terms. As will be illustrated in the experiments, HOMI is quite robust to those hyper-parameters. It is also worth pointing out that HOMI integrates high-order correlations exploitation and model prediction into a joint model via the graph Laplacian regularization term. It is able to simultaneously enhance those two processes via the joint learning.

\subsection{Prediction}
For an unseen instance $\mathbf{x}$, the discriminative function $\mathbf{g}$ of HOMI is obtained by
\begin{equation}
\mathbf{g}(
\mathbf{x}) = (\mathbf{xW}+\mathbf{z}^\mathsf{T})\mathbf{B} + \mathbf{t}^\mathsf{T},
\label{gx}
\end{equation}
and the predictive label set is obtained by
\begin{equation}
    \mathbf{y}_{pre} = f(\mathbf{x})= \{c_i |\mathbf{g}_i(\mathbf{x})>0.5, 1\le i\le l\},
    \label{fx}
\end{equation}
where $\mathbf{g}_i(\mathbf{x})$ is the $i$th element of $\mathbf{g}(\mathbf{x})$.

\subsection{Numerical Solution to Eq. (\ref{loss})}
The problem in Eq. (\ref{loss}) is not convex in all the variables together, but it is convex to each variable with the remaining variables fixed. Therefore, we solved it by the following alternating optimization procedure.

\noindent
\textbf{Update W} With fixed $\mathbf{B}$, $\mathbf{z}$ and $\mathbf{t}$, Eq. (\ref{loss}) becomes:
\begin{equation}
    \min_{\mathbf{W}}\|\mathbf{Y}-\mathbf{XW}-\mathbf{{1_n}z}^\mathsf{T}\|^2_F+ \gamma{\rm tr}(\mathbf{K}^\mathsf{T}\mathbf{LK})+ \lambda\|\mathbf{W}\|^2_F.
    \label{minw}
\end{equation}
Taking the gradient of Eq. (\ref{minw}) w.r.t. $\mathbf{W}$, we have
\begin{equation}
\begin{split}
     \nabla_{\mathbf{W}} = \mathbf{X}^\mathsf{T}(\mathbf{XW}+\mathbf{1}_n\mathbf{ z}^\mathsf{T}-\mathbf{Y})+\lambda \mathbf{W} \\ +\gamma \mathbf{X}^\mathsf{T}\mathbf{L}(\mathbf{XWB} + \mathbf{1}_n\mathbf{ z}^\mathsf{T}\mathbf{B}+\mathbf{1}_n\mathbf{t}^\mathsf{T})\mathbf{B}^\mathsf{T}.
\end{split}
\label{deltaw}
\end{equation}
The optimal solution of Eq. (\ref{minw}) is achieved when $\nabla_{\mathbf{W}} = 0$, and accordingly we have the following Sylvester equation

\begin{equation}
\mathbf{AW}+\mathbf{WE} = \mathbf{Q}\vspace{1ex}
\label{W=}
\end{equation}
where $\mathbf{A} = \frac{1}{\gamma}(\mathbf{X}^\mathsf{T}\mathbf{LX})^{-1}(\mathbf{X}^\mathsf{T}\mathbf{X}+\lambda \mathbf{I})$, $\mathbf{E} = \mathbf{BB}^\mathsf{T}\vspace{1ex}$, $\mathbf{Q} = \frac{1}{\gamma}(\mathbf{X}^\mathsf{T}\mathbf{LX})^{-1}(\mathbf{X}^\mathsf{T}\mathbf{Y} - \mathbf{X}^\mathsf{T} \mathbf{1}_n\mathbf{ z}^\mathsf{T}-\vspace{1ex} \gamma \mathbf{X}^\mathsf{T}\mathbf{L 1_n t}^\mathsf{T}\mathbf{B}^\mathsf{T}-\gamma \mathbf{X}^\mathsf{T}\mathbf{L 1_n z}^\mathsf{T}\mathbf{BB}^\mathsf{T})\vspace{1ex}$,  which can be efficiently solved according to \cite{1102170}.

\noindent
\textbf{Update B} Fixing $\mathbf{W}$,$\mathbf{z}$ and $\mathbf{t}$, the $\mathbf{B}$-subproblem becomes
$$\min_{\mathbf{B}}\beta\|\mathbf{Y}-\mathbf{YB}-\mathbf{1}_n\mathbf{t}^\mathsf{T}\|_F^2 +
        \gamma {\rm tr}(\mathbf{K}^\mathsf{T}\mathbf{LK})+ \lambda\|\mathbf{B}\|^2_F,$$
which is a quadratic optimization problem, and the solution is obtained by setting its derivative to $0$, i.e.,
\begin{equation}
    \begin{split}
    \mathbf{B} = &(\beta \mathbf{Y}^\mathsf{T}\mathbf{Y} + \lambda \mathbf{I} + \gamma (\mathbf{W}^\mathsf{T}\mathbf{X}^\mathsf{T}+\mathbf{z1_n}^\mathsf{T}) \mathbf{L} (\mathbf{XW}+\mathbf{1}_n\mathbf{ z}^\mathsf{T}))^{-1}\\& (\beta \mathbf{Y}^\mathsf{T}\mathbf{Y}- \beta \mathbf{Y}^\mathsf{T} \mathbf{1}_n\mathbf{t}^\mathsf{T} -\gamma (\mathbf{W}^\mathsf{T}\mathbf{X}^\mathsf{T} + \mathbf{z1_n}^\mathsf{T})\mathbf{L1_nt}^\mathsf{T}).
    \end{split}
    \label{B=}
\end{equation}

\noindent
\textbf{Update z} With fixed $\mathbf{W}$, $\mathbf{B}$ and $\mathbf{t}$, the $\mathbf{z}$-subproblem is re-written as
$$\min_{\mathbf{z}} \|\mathbf{Y}-\mathbf{XW}-\mathbf{1}_n\mathbf{ z}^\mathsf{T}\|_F^2 +
        \gamma{\rm tr}(\mathbf{K}^\mathsf{T}\mathbf{LK})+ \lambda\|\mathbf{z}\|^2_2,$$
which is also a quadratic optimization problem, and the optimal solution is achieved when the derivative approaches zero, i.e., 
\begin{equation}
    \begin{split}
    &\mathbf{z} = (({n} + \lambda)\mathbf{I} + \gamma \mathbf{1}_n\mathbf{}^\mathsf{T}\mathbf{L}\mathbf{1}_n\mathbf{BB}^\mathsf{T})^{-1} (\mathbf{Y}^\mathsf{T}\mathbf{{1_n}} - \mathbf{W}^\mathsf{T}\mathbf{X}^\mathsf{T}\mathbf{{1_n}}  \\ & -\gamma \mathbf{B}(\mathbf{B}^\mathsf{T}\mathbf{ W}^\mathsf{T}\mathbf{X}^\mathsf{T}+\mathbf{t1_n}^\mathsf{T}) \mathbf{L}\mathbf{{1_n}}).
    \end{split}
    \label{z=}
\end{equation}

\noindent
\textbf{Update t} With other variables fixed, the $\mathbf{t}$-subproblem is reformulated as
$$\min_{\mathbf{t}}\beta\|\mathbf{Y}-\mathbf{YB}-\mathbf{{1_n}t}^\mathsf{T}\|_F^2 +
        \gamma {\rm tr}(\mathbf{K}^\mathsf{T}\mathbf{LK})+ \lambda\|\mathbf{t}\|^2_2,$$
which is also a quadratic optimization problem, and the solution of it is achieved when $\nabla_{\mathbf{t}} = 0$, i.e., 
\begin{equation}
    \begin{split}
    \mathbf{t} = & \frac{1}{(\beta  n + \gamma \mathbf{1}_n\mathbf{}^\mathsf{T}\mathbf{L}\mathbf{1}_n\mathbf{} + \lambda)} (\beta \mathbf{Y}^\mathsf{T}\mathbf{1}_n\mathbf{}  \\& - \beta \mathbf{B}^\mathsf{T}\mathbf{Y}^\mathsf{T}\mathbf{1}_n\mathbf{} - \gamma \mathbf{B}^\mathsf{T}(\mathbf{W}^\mathsf{T}\mathbf{X}^\mathsf{T}+\mathbf{z1_n}^\mathsf{T})\mathbf{L}\mathbf{1}_n\mathbf{}).
    \end{split}
    \label{t=}
\end{equation}

In summary, HOMI first randomly initialize $\mathbf{W, B, z, t}$, and then iteratively and alternatively update these four variables. The iteration stops when the difference between two consecutive loss is less than 0.001. Finally the label set $\mathbf{y}_{pre}$ for an unseen $\mathbf{x}$ is predicted according to Eq. (\ref{fx}). The whole schedule is concluded in algorithm 1.

\begin{algorithm}[htb]
  \caption{HOMI.}
  \label{alg:Framwork}
  \begin{algorithmic}[1]
    \Require
        Training data set $\mathcal{D}$; $s$; $\beta$; $\gamma$; $\lambda$, max iteration number $iter$; an unseen instance $\mathbf{x}$.
    \Ensure
      Predicted label set $\mathbf{y}_{pre}$ for $\mathbf{x}$.
    \State Initialize $\mathbf{W, B, z, t}$ to $\mathbf0$;
    \State Calculate $\mathbf{L}$ by Eqs. (\ref{sij})-(\ref{geometric});
    \State \textbf{Repeat:}
    \State Update $\mathbf{W}$ according to Eq. (\ref{W=});
    \State Update $\mathbf{B}$ according to Eq. (\ref{B=});
    \State Update $\mathbf{z}$ according to Eq. (\ref{z=});
    \State Update $\mathbf{t}$ according to Eq. (\ref{t=});
    \State \textbf{If} number of iteration $\ge$ $iter$: \textbf{break}
    \State \textbf{Until} Convergence;\\
    \Return $\mathbf{y}_{pre}$ according to Eq. (\ref{fx}).
  \end{algorithmic}
\end{algorithm}

\subsection{Complexity Analysis}
HOMI iteratively solves four optimization problems. To solve $\mathbf{W}$, HOMI needs to solve a Sylvester equation, which can be computed in $O(\max (m^3, l^3, mn^2, m^2n, mnl))$; the rest three optimization problems are all quadratic optimization problems with the complexity of $O(\max(l^2n, lmn, l^3))$, $O(\max(l^3, lmn, n^2))$ and $O(\max(l^3, lmn, n^2))$. In summary, the overall complexity of HOMI is $O(\max(m^3,l^3,mn^2,m^2n,mnl))$ in one iteration.

\subsection{Convergence Analysis}
The proposed numerical solution in Algorithm 1 is guaranteed to converge theoretically. 
Specifically, define the optimal function in Eq. (\ref{loss}) as $F$, the problem in Eq. (\ref{loss}) is not convex in all the variables together, but it is convex to each variable with the remaining variables fixed. Therefore, we solved it by an alternating optimization procedure. Specifically, we transform the original problem into four subproblems, where each subproblems can be solved efficiently. We need to minimize $F(\mathbf{W,B,z,t})$ with four variables $\mathbf{W,B,z}$ and $\mathbf{t}$. We transform the original problem into four subproblems $\min_\mathbf{W} F(\mathbf{W,B,z,t})$, $\min_\mathbf{B} F(\mathbf{W,B,z,t})$, $\min_\mathbf{z} F(\mathbf{W,B,z,t})$ and $\min_\mathbf{t} F(\mathbf{W,B,z,t})$ and solve them alternatively and iteratively. When solving the $\mathbf{W}$-subproblem $\min_\mathbf{W} F(\mathbf{W}_{k-1},\mathbf{B}_{k-1},\mathbf{z}_{k-1},\mathbf{t}_{k-1})$ at the $k$th iteration, the variables $\mathbf{B}_{k-1},\mathbf{z}_{k-1},\mathbf{t}_{k-1}$ are fixed, and we try to find optimal $\mathbf{W}_{k}$ to minimize the corresponding function value. It is obvious that $F(\mathbf{W}_{k-1},
\mathbf{B}_{k-1},\mathbf{z}_{k-1},\mathbf{t}_{k-1}) \geq F(\mathbf{W}_{k},
\mathbf{B}_{k-1},\mathbf{z}_{k-1},\mathbf{t}_{k-1})$. Similarly, when solving the $\mathbf{B}$-subproblem $\min_\mathbf{B} F(\mathbf{W}_{k},
\mathbf{B}_{k-1},\mathbf{z}_{k-1},\mathbf{t}_{k-1})$ at the $k$th iteration, the variables $\mathbf{W}_{k},\mathbf{z}_{k-1},\mathbf{t}_{k-1}$ are fixed, and $F(\mathbf{W}_{k},
\mathbf{B}_{k-1}, \mathbf{z}_{k-1},\mathbf{t}_{k-1}) \geq F(\mathbf{W}_{k},
\mathbf{B}_{k},\mathbf{z}_{k-1},\mathbf{t}_{k-1})$, and the same to $\mathbf{z}_k$ and $\mathbf{t}_k$. Therefore, we can get $F(\mathbf{W}_{k-1},
\mathbf{B}_{k-1},\mathbf{z}_{k-1},\mathbf{t}_{k-1}) \geq F(\mathbf{W}_{k},
\mathbf{B}_{k},\mathbf{z}_{k},\mathbf{t}_{k})$, i.e., in each iteration, the value of the loss function is decreased. As the loss function has a lower bound ($F\geq 0$), the above alternating algorithm will surely be converged.

\section{Experiments}
\begin{table*}[!htb]\scriptsize
\setlength{\tabcolsep}{6.7mm}\renewcommand\arraystretch{0.8} 
     \caption{Characteristics of the experimental data sets.}
    \centering
   
    \begin{tabular}{c c c c c c c c}\hline\hline
         Data set &  $n$ & $m$ & $l$ & $LCard(\mathcal{D})$ & $LDen(\mathcal{D})$ & $DL(\mathcal{D})$ & Domain \\\hline
         
          mediamill & 43097 & 120 & 101 & 4.376 & 0.043 & 6555 & video\\
          
         CAL500 & 502 & 68 & 174 & 26.044 & 0.150 & 502 & music\\
         
         emotions & 593 & 72 & 6 & 1.869 & 0.311 & 27 & music\\
          
         enron & 1702 & 1001 & 53 & 3.378 & 0.064 & 753 & text\\
         
         bibtex & 7395 & 1836 & 159 & 2.402 & 0.015 & 2856 & text\\
         
         delicious & 16105 & 500 & 983 & 19.020 & 0.019 & 15806 & text\\
         
         language log & 1460 & 1004 & 75 & 1.180 & 0.016 & 286 & text\\
         
         birds & 645 & 260 & 19 & 1.014 & 0.053 & 133 & audio\\
         yeast & 2417 & 103 &  14 & 4.237 & 0.303 & 198 & biology\\
         scene & 2407 & 294 & 6 & 1.074 & 0.179 & 15 & images\\
         corel5k & 5000 & 499 & 374 & 3.52 & 0.009 & 3175 & images\\
         corel16k & 13766 & 500 & 153 & 2.859& 0.019& 4803 & images\\
         \hline\hline
    \end{tabular}
   
    \label{tab:chara}
\end{table*}

\begin{table*}[!htb]\tiny
\setlength{\tabcolsep}{0.85mm}
\renewcommand\arraystretch{1.2} 
    \centering
    
     \caption{The comparison of different methods on all the data sets with respect to Hamming Loss. $\bullet$ / $\circ$ indicates whether HOMI is superior or same / inferior to the compared algorithm.}
    \begin{tabular}{c l l l l l l l l l l l l}\hline\hline
          \multicolumn{13}{c}{Hamming loss $\downarrow$}\\\hline
         Approach &\multicolumn{1}{c}{mediamill}& \multicolumn{1}{c}{CAL500} & \multicolumn{1}{c}{emotions} & \multicolumn{1}{c}{enron} &\multicolumn{1}{c}{bibtex} &\multicolumn{1}{c}{delicious} &\multicolumn{1}{c}{language log} &\multicolumn{1}{c}{birds} & \multicolumn{1}{c}{yeast} & \multicolumn{1}{c}{scene} & \multicolumn{1}{c}{corel5k}&\multicolumn{1}{c}{corel16k1} \\\hline
         ECC &0.031 $\pm$ 0.000 $\bullet$ & 0.147 $\pm$ 0.002 $\bullet$ & 0.226 $\pm$ 0.011 $\bullet$ &0.050 $\pm$ 0.001 $\bullet$ & 0.013 $\pm$ 0.000 $\bullet$  & 0.019 $\pm$ 0.000 $\bullet$ & 0.015 $\pm$ 0.001 $\bullet$ &0.050 $\pm$ 0.004 $\circ$  & 0.211 $\pm$ 0.008 $\bullet$ &0.122 $\pm$ 0.003 $\bullet$ & 0.010 $\pm$ 0.000 $\bullet$& 0.020 $\pm$ 0.000 $\bullet$\\

         BR &0.029 $\pm$ 0.000 $\bullet$ & 0.165 $\pm$ 0.005 $\bullet$ & 0.207 $\pm$ 0.021 $\bullet$ & 0.061 $\pm$ 0.002 $\bullet$ & 0.016 $\pm$ 0.000 $\bullet$ &0.018 $\pm$ 0.000 $\bullet$ & 0.019 $\pm$ 0.001 $\bullet$ & 0.070 $\pm$ 0.010 $\bullet$ & 0.202 $\pm$ 0.006 $\bullet$ & 0.131 $\pm$ 0.005 $\bullet$ & 0.011 $\pm$ 0.000 $\bullet$ & 0.019 $\pm$ 0.000 $\bullet$ \\
         ML-KNN &0.029 $\pm$ 0.001 $\bullet$ & 0.147 $\pm$ 0.003 $\bullet$ & 0.200 $\pm$ 0.007 $\circ$ & 0.055 $\pm$ 0.002 $\bullet$ & 0.015 $\pm$ 0.000 $\bullet$ &0.019 $\pm$ 0.000 $\bullet$ &0.051 $\pm$ 0.003 $\bullet$& 0.051 $\pm$ 0.006 $\bullet$ & 0.199 $\pm$ 0.008 $\circ$ & 0.086 $\pm$ 0.005 $\circ$ & 0.010 $\pm$ 0.000 $\bullet$ & 0.020 $\pm$ 0.000 $\bullet$ \\
         WRAP &0.030 $\pm$ 0.000 $\bullet$ & 0.136 $\pm$ 0.002 $\bullet$ & 0.200 $\pm$ 0.007 $\circ$ & 0.047 $\pm$ 0.001 $\circ$ & 0.012 $\pm$ 0.000 $\bullet$ &0.019 $\pm$ 0.000 $\bullet$ &0.015 $\pm$ 0.001 $\bullet$& 0.045 $\pm$ 0.048  & 0.199 $\pm$ 0.005 $\circ$ & 0.110 $\pm$ 0.006 $\circ$ & 0.009 $\pm$ 0.000 $\bullet$ & 0.019 $\pm$ 0.001 $\bullet$\\
         MLSF &0.030 $\pm$ 0.000 $\bullet$ & 0.138 $\pm$ 0.004 $\bullet$ & 0.224 $\pm$ 0.021 $\bullet$ & 0.051 $\pm$ 0.001 $\bullet$ & 0.012 $\pm$ 0.000 $\bullet$ &0.028 $\pm$ 0.000 $\bullet$ & 0.015 $\pm$ 0.001 $\bullet$& 0.050 $\pm$ 0.003 $\circ$ & 0.210 $\pm$ 0.005 $\bullet$ & 0.128 $\pm$ 0.007 $\bullet$ & 0.009 $\pm$ 0.000 $\bullet$ & 0.019 $\pm$ 0.000 $\bullet$\\
         LFLC & 0.030 $\pm$ 0.000 $\bullet$ & 0.136 $\pm$ 0.004 $\bullet$ & 0.197 $\pm$ 0.004 $\circ$ & 0.046 $\pm$ 0.001 $\circ$ & 0.012 $\pm$ 0.000 $\bullet$ &0.024 $\pm$ 0.000 $\bullet$ & 0.015 $\pm$ 0.001 $\bullet$ & 0.045 $\pm$ 0.003 $\circ$ & 0.197 $\pm$ 0.004 $\circ$ & 0.107 $\pm$ 0.005 $\circ$ & 0.009 $\pm$ 0.000 $\bullet$ & 0.019 $\pm$ 0.000 $\bullet$\\
         BILAS &0.029 $\pm$ 0.001 $\bullet$ & 0.138 $\pm$ 0.000 $\bullet$ & 0.233 $\pm$ 0.000 $\bullet$ & 0.045 $\pm$ 0.000 $\circ$ & 0.011 $\pm$ 0.000 $\bullet$ &0.044 $\pm$ 0.000 $\bullet$ & 0.018 $\pm$ 0.002 $\bullet$& 0.052 $\pm$ 0.000 $\bullet$ & 0.198 $\pm$ 0.000 $\circ$ & 0.097 $\pm$ 0.000 $\circ$ & 0.009 $\pm$ 0.000 $\bullet$ & 0.018 $\pm$ 0.000 $\bullet$\\
         GLOCAL & 0.045 $\pm$ 0.000 $\bullet$ & 0.143 $\pm$ 0.008 $\bullet$& 0.323 $\pm$ 0.020 $\bullet$ & 0.064 $\pm$ 0.001 $\bullet$ &0.014 $\pm$ 0.000 $\bullet$ &0.022 $\pm$ 0.002 $\bullet$& 0.018 $\pm$ 0.000 $\bullet$ & 0.097 $\pm$ 0.004 $\bullet$ &0.300 $\pm$ 0.003 $\bullet$ &0.179 $\pm$ 0.000 $\bullet$ & 0.009 $\pm$ 0.000 $\bullet$ & 0.019 $\pm$ 0.001 $\bullet$\\
         
         ML-LRC & 0.042 $\pm$ 0.000 $\bullet$ & 0.150 $\pm$ 0.003 $\bullet$ & 0.273 $\pm$ 0.006 $\bullet$ & 0.050 $\pm$ 0.013 $\bullet$ & 0.010 $\pm$ 0.000 $\bullet$ &0.019 $\pm$ 0.000 $\bullet$ & 0.072 $\pm$ 0.004 $\bullet$  & 0.055 $\pm$ 0.006 $\bullet$ & 0.296 $\pm$ 0.003 $\bullet$ & 0.138 $\pm$ 0.002 $\bullet$ & 0.009 $\pm$ 0.000 $\bullet$ & 0.019 $\pm$ 0.000 $\bullet$\\

         CLML & 0.049 $\pm$ 0.001 $\bullet$ & 0.192 $\pm$ 0.006 $\bullet$ & 0.315 $\pm$ 0.011 $\bullet$ & 0.051 $\pm$ 0.002 $\bullet$ & 0.016 $\pm$ 0.000 $\bullet$ & 0.027 $\pm$ 0.001 $\bullet$ & 0.020 $\pm$ 0.002 $\bullet$ & 0.046 $\pm$ 0.005 $\circ$ & 0.304 $\pm$ 0.005 $\bullet$ & 0.179 $\pm$ 0.003 $\bullet$ & 0.015 $\pm$ 0.000 $\bullet$ & 0.019 $\pm$ 0.000 $\bullet$

         \\\hline

         HOMI & 0.029 $\pm$ 0.000 & 0.135 $\pm$ 0.001  & 0.205 $\pm$ 0.023  & 0.049 $\pm$ 0.012 & 0.009 $\pm$ 0.000 & 0.018 $\pm$ 0.000 & 0.015 $\pm$ 0.000 & 0.051 $\pm$ 0.006 & 0.200 $\pm$ 0.005 & 0.112 $\pm$ 0.005 & 0.009 $\pm$ 0.000 & 0.018 $\pm$ 0.000\\\hline\hline

    \end{tabular}
   
    \label{tab:hamming}
\end{table*}

\begin{table*}[!htb]\tiny
\setlength{\tabcolsep}{0.85mm}
\renewcommand\arraystretch{1.2} 
    \centering
     \caption{The comparison of different methods on all the data sets with respect to Ranking Loss. $\bullet$ / $\circ$ indicates whether HOMI is superior or same / inferior to the compared algorithm.}
    \begin{tabular}{c l l l l l l l l l l l l}\hline\hline
          \multicolumn{13}{c}{Ranking loss $\downarrow$}\\\hline
         Approach &\multicolumn{1}{c}{mediamill}& \multicolumn{1}{c}{CAL500} & \multicolumn{1}{c}{emotions} & \multicolumn{1}{c}{enron} &\multicolumn{1}{c}{bibtex}&\multicolumn{1}{c}{delicious}&\multicolumn{1}{c}{language log} & \multicolumn{1}{c}{birds} & \multicolumn{1}{c}{yeast} & \multicolumn{1}{c}{scene} & \multicolumn{1}{c}{corel5k}& \multicolumn{1}{c}{corel16k1}\\\hline
        
         ECC & 0.092 $\pm$ 0.002 $\bullet$ & 0.214 $\pm$ 0.007 $\bullet$ & 0.183 $\pm$ 0.017 $\bullet$ & 0.088 $\pm$ 0.005 $\bullet$ & 0.088 $\pm$ 0.002 $\bullet$ &0.141 $\pm$ 0.002 $\circ$ &0.112 $\pm$ 0.011 $\circ$& 0.097 $\pm$ 0.008 $\circ$ & 0.202 $\pm$ 0.009 $\bullet$ & 0.138 $\pm$ 0.004 $\bullet$ & 0.148 $\pm$ 0.003 $\bullet$ & 0.180 $\pm$ 0.002 $\bullet$\\

         BR &0.036 $\pm$ 0.000 $\circ$ & 0.180 $\pm$ 0.006 $\bullet$ & 0.162 $\pm$ 0.014 $\circ$ & 0.085 $\pm$ 0.004 $\bullet$ & 0.086 $\pm$ 0.003 $\bullet$ & 0.121 $\pm$ 0.001 $\circ$&0.113 $\pm$ 0.004 $\circ$& 0.104 $\pm$ 0.007 $\circ$ & 0.174 $\pm$ 0.011 $\bullet$ & 0.115 $\pm$ 0.013 $\bullet$ & 0.123 $\pm$ 0.004 $\circ$ & 0.162 $\pm$ 0.002 $\bullet$\\
         ML-KNN &0.046 $\pm$ 0.000 $\bullet$ & 0.220 $\pm$ 0.005 $\bullet$ & 0.214 $\pm$ 0.009 $\bullet$ & 0.111 $\pm$ 0.003 $\bullet$ & 0.252 $\pm$ 0.006 $\bullet$ & 0.166 $\pm$ 0.002 & 0.156 $\pm$ 0.006 $\bullet$ & 0.125 $\pm$ 0.023 $\bullet$ & 0.221 $\pm$ 0.007 $\bullet$ & 0.105 $\pm$ 0.007 $\circ$ & 0.149 $\pm$ 0.004 $\bullet$ & 0.196 $\pm$ 0.001 $\bullet$\\
         WRAP &0.048 $\pm$ 0.001 $\bullet$ & 0.175 $\pm$ 0.006 $\circ$ & 0.157 $\pm$ 0.028 $\circ$ & 0.080 $\pm$ 0.006 $\bullet$ & 0.102 $\pm$ 0.003 $\bullet$ &0.118 $\pm$ 0.002 $\circ$&0.157 $\pm$ 0.015 $\bullet$& 0.088 $\pm$ 0.011 $\circ$ & 0.170 $\pm$ 0.004 $\circ$ & 0.103 $\pm$ 0.006 $\circ$ & 0.142 $\pm$ 0.004 $\circ$ & 0.145 $\pm$ 0.001 $\circ$\\
         MLSF &0.054 $\pm$ 0.005 $\bullet$ & 0.204 $\pm$ 0.004 $\bullet$ & 0.181 $\pm$ 0.014 $\bullet$ & 0.084 $\pm$ 0.005 $\bullet$ & 0.092 $\pm$ 0.015 $\bullet$ &0.127 $\pm$ 0.002 $\circ$&0.118 $\pm$ 0.019 $\circ$& 0.107 $\pm$ 0.027 $\circ$ & 0.201 $\pm$ 0.008 $\bullet$ & 0.122 $\pm$ 0.010 $\bullet$ & 0.143 $\pm$ 0.004 $\circ$ & 0.183 $\pm$ 0.006 $\bullet$\\
         LFLC &0.047 $\pm$ 0.001 $\bullet$ & 0.177 $\pm$ 0.004 $\circ$ & 0.170 $\pm$ 0.008 $\bullet$ & 0.080 $\pm$ 0.007 $\bullet$ & 0.098 $\pm$ 0.003 $\bullet$ &0.169 $\pm$ 0.002 $\bullet$&0.143 $\pm$ 0.013 $\circ$ & 0.086 $\pm$ 0.010 $\circ$ & 0.171 $\pm$ 0.009 $\bullet$ & 0.084 $\pm$ 0.006 $\circ$ & 0.162 $\pm$ 0.008 $\bullet$ & 0.171 $\pm$ 0.004 $\bullet$\\         
         BILAS &0.048 $\pm$ 0.001 $\bullet$ & 0.181 $\pm$ 0.000 $\bullet$ & 0.217 $\pm$ 0.000 $\bullet$ & 0.065 $\pm$ 0.000 $\circ$ & 0.077 $\pm$ 0.003 $\bullet$ &0.129 $\pm$ 0.001 $\circ$ &0.166 $\pm$ 0.007 $\bullet$& 0.355 $\pm$ 0.000 $\bullet$ & 0.162 $\pm$ 0.000 $\circ$ & 0.075 $\pm$ 0.000 $\circ$ & 0.201 $\pm$ 0.004 $\bullet$ & 0.172 $\pm$ 0.004 $\bullet$  \\
         GLOCAL & 0.055 $\pm$ 0.009 $\bullet$ & 0.182 $\pm$ 0.016 $\bullet$ &0.260 $\pm$ 0.009 $\bullet$ & 0.039 $\pm$ 0.004 $\circ$ & 0.093 $\pm$ 0.010 $\bullet$ &0.243 $\pm$ 0.002 $\bullet$&0.297 $\pm$ 0.004 $\bullet$& 0.293 $\pm$ 0.021 $\bullet$ & 0.362 $\pm$ 0.088 $\bullet$ & 0.177 $\pm$ 0.004 $\bullet$ & 0.227 $\pm$ 0.010 $\bullet$ & 0.189 $\pm$ 0.005 $\bullet$ \\ 
         
         ML-LRC & 0.206 $\pm$ 0.002 $\bullet$ & 0.485 $\pm$ 0.013 $\bullet$ & 0.190 $\pm$ 0.023 $\bullet$ & 0.106 $\pm$ 0.024 $\bullet$ & 0.257 $\pm$ 0.002 $\bullet$ &0.238 $\pm$ 0.004 $\bullet$&0.336 $\pm$ 0.015 $\bullet$& 0.145 $\pm$ 0.021 $\bullet$ & 0.353 $\pm$ 0.015 $\bullet$ & 0.110 $\pm$ 0.011 $\bullet$ & 0.314 $\pm$ 0.002 $\bullet$ & 0.291 $\pm$ 0.003 $\bullet$\\

         CLML & 0.449 $\pm$ 0.009 $\bullet$ & 0.180 $\pm$ 0.005 $\bullet$ & 0.197 $\pm$ 0.014 $\bullet$ & 0.082 $\pm$ 0.004 $\bullet$ & 0.071 $\pm$ 0.003 $\bullet$ & 0.135 $\pm$ 0.002 $\circ$ & 0.128 $\pm$ 0.004 $\circ$ & 0.099 $\pm$ 0.022 $\circ$ & 0.356 $\pm$ 0.012 $\bullet$ & 0.111 $\pm$ 0.014 $\bullet$ & 0.162 $\pm$ 0.004 $\bullet$ & 0.282 $\pm$ 0.005 $\bullet$
         
         \\\hline
         HOMI & 0.043 $\pm$ 0.001 & 0.179 $\pm$ 0.005  & 0.169 $\pm$ 0.026 & 0.078 $\pm$ 0.015 & 0.024 $\pm$ 0.000 & 0.170 $\pm$ 0.003 & 0.147 $\pm$ 0.002& 0.110 $\pm$ 0.016 & 0.171 $\pm$ 0.006 & 0.108 $\pm$ 0.012 & 0.145 $\pm$ 0.002 & 0.155 $\pm$ 0.003 \\\hline\hline
    
    \end{tabular}
  
    \label{tab:rank}
\end{table*}

\begin{table*}[!htb]\tiny
\setlength{\tabcolsep}{0.85mm}
\renewcommand\arraystretch{1.2} 
    \centering
      \caption{The comparison of different methods on all the data sets with respect to One-error. $\bullet$ / $\circ$ indicates whether HOMI is superior or same / inferior to the compared algorithm.}
    \begin{tabular}{c l l l l l l l l l l l l}\hline\hline
          \multicolumn{13}{c}{One-error $\downarrow$}\\\hline
         Approach &\multicolumn{1}{c}{mediamill} &  \multicolumn{1}{c}{CAL500} & \multicolumn{1}{c}{emotions} & \multicolumn{1}{c}{enron} &
         \multicolumn{1}{c}{bibtex} &\multicolumn{1}{c}{delicious}&\multicolumn{1}{c}{language log}&
         \multicolumn{1}{c}{birds} & \multicolumn{1}{c}{yeast} & \multicolumn{1}{c}{scene} & \multicolumn{1}{c}{corel5k} &
         \multicolumn{1}{c}{corel16k1}\\\hline
         
         ECC &0.176 $\pm$ 0.005 $\bullet$ & 0.299 $\pm$ 0.042 $\bullet$ & 0.312 $\pm$ 0.033 $\bullet$ & 0.267 $\pm$ 0.026 $\bullet$ & 0.402 $\pm$ 0.015 $\bullet$ &0.467 $\pm$ 0.010 $\circ$& 0.750 $\pm$ 0.016 $\circ$ & 0.712 $\pm$ 0.035 $\bullet$ & 0.262 $\pm$ 0.023 $\bullet$ & 0.338 $\pm$ 0.018 $\bullet$ & 0.757 $\pm$ 0.014 $\bullet$ & 0.693 $\pm$ 0.003 $\bullet$\\

         BR &0.162 $\pm$ 0.006 $\circ$ & 0.119 $\pm$ 0.045 $\bullet $ & 0.268 $\pm$ 0.024 $\circ$ & 0.271 $\pm$ 0.015 $\bullet$ & 0.399 $\pm$ 0.011 $\bullet$ &0.343 $\pm$ 0.008 $\circ$&0.750 $\pm$ 0.008 $\circ$& 0.712 $\pm$ 0.045 $\bullet$ & 0.233 $\pm$ 0.033  & 0.317 $\pm$ 0.030 $\bullet$ & 0.667 $\pm$ 0.017 $\circ$ & 0.738 $\pm$ 0.010  $\bullet$\\
         ML-KNN &0.182 $\pm$ 0.004 $\bullet$ & 0.158 $\pm$ 0.014 $\bullet$ & 0.372 $\pm$ 0.004 $\bullet$ & 0.468 $\pm$ 0.018 $\bullet$ & 0.837 $\pm$ 0.006 $\bullet$ & 0.573 $\pm$ 0.012 $\bullet$&0.905 $\pm$ 0.016 $\bullet$& 0.837 $\pm$ 0.029 $\bullet$ & 0.254 $\pm$ 0.023 $\bullet$ & 0.317 $\pm$ 0.013 $\bullet$ & 0.791 $\pm$ 0.009 $\bullet$ & 0.805 $\pm$ 0.007 $\bullet$\\
         WRAP &0.156 $\pm$ 0.003 $\circ$ & 0.113 $\pm$ 0.033 $\bullet $ & 0.265 $\pm$ 0.025 $\circ$ & 0.220 $\pm$ 0.021 $\bullet$  & 0.363 $\pm$ 0.012 $\bullet$ &0.340 $\pm$ 0.010 $\circ$& 0.730 $\pm$ 0.032 $\circ$& 0.660 $\pm$ 0.031 $\circ$ & 0.223 $\pm$ 0.008 $\circ$ & 0.272 $\pm$ 0.008 $\circ$ & 0.621 $\pm$ 0.032 $\circ$ & 0.635 $\pm$ 0.005 $\circ$\\
         MLSF &0.193 $\pm$ 0.008 $\bullet$ & 0.128 $\pm$ 0.040 $\bullet$ & 0.319 $\pm$ 0.021 $\bullet$ & 0.283 $\pm$ 0.028 $\bullet$ & 0.402 $\pm$ 0.013 $\bullet$ & 0.358 $\pm$ 0.009 $\circ$ &0.755 $\pm$ 0.012 $\circ$& 0.701 $\pm$ 0.055 $\bullet$ & 0.261 $\pm$ 0.022 $\bullet$ & 0.345 $\pm$ 0.010 $\bullet$ & 0.667 $\pm$ 0.010 $\circ$ & 0.743 $\pm$ 0.019 $\bullet$\\
         LFLC & 0.158 $\pm$ 0.004 $\circ$ & 0.120 $\pm$ 0.022 $\bullet $ & 0.227 $\pm$ 0.031 $\circ$ & 0.231 $\pm$ 0.009 $\bullet$ & 0.351 $\pm$ 0.007 $\bullet$ & 0.343 $\pm$ 0.008 $\circ$ &0.721 $\pm$ 0.025 $\circ$ & 0.656 $\pm$ 0.037 $\circ$ & 0.225 $\pm$ 0.025 $\circ$ & 0.250 $\pm$ 0.023 $\circ$ & 0.631 $\pm$ 0.018 $\circ$ & 0.645 $\pm$ 0.010 $\circ$\\
         BILAS &0.166 $\pm$ 0.003 $\bullet$ & 0.115 $\pm$ 0.000 $\bullet$ & 0.357 $\pm$ 0.000 $\bullet$ & 0.233 $\pm$ 0.000 $\bullet$ & 0.361 $\pm$ 0.006 $\bullet$ &0.543 $\pm$ 0.013 $\circ$&0.877 $\pm$ 0.015 $\bullet$& 0.868 $\pm$ 0.000 $\bullet$ & 0.223 $\pm$ 0.000 $\circ$ & 0.229 $\pm$ 0.000 $\circ$ & 0.776 $\pm$ 0.000 $\bullet$ & 0.724 $\pm$ 0.003 $\bullet$  \\
         GLOCAL & 0.171 $\pm$ 0.005 $\bullet$ & 0.118 $\pm$ 0.001 $\bullet$ & 0.432 $\pm$ 0.000 $\bullet$ & 0.077 $\pm$ 0.010 $\circ$ & 0.433 $\pm$ 0.031 $\bullet$ &0.672 $\pm$ 0.003 $\bullet$&0.881 $\pm$ 0.004 $\bullet$& 0.684 $\pm$ 0.008 $\circ$ & 0.319 $\pm$ 0.040 $\bullet$  & 0.479 $\pm$ 0.021 $\bullet$ & 0.670 $\pm$ 0.220 $\circ$ & 0.644 $\pm$ 0.070 $\circ$\\
         
         ML-LRC & 0.322 $\pm$ 0.008 $\bullet$ & 0.603 $\pm$ 0.048 $\bullet$ & 0.291 $\pm$ 0.036 $\bullet$ & 0.141 $\pm$ 0.034 $\circ$ & 0.141 $\pm$ 0.007 $\bullet$ &0.403 $\pm$ 0.018 $\circ$&0.879 $\pm$ 0.017 $\bullet$& 0.719 $\pm$ 0.031 $\bullet$ & 0.368 $\pm$ 0.009 $\bullet$ & 0.285 $\pm$ 0.022 $\bullet$ & 0.683 $\pm$ 0.015 $\circ$ & 0.674 $\pm$ 0.012 $\bullet$\\

         CLML & 0.305 $\pm$ 0.011 $\bullet$ & 0.115 $\pm$ 0.026 $\bullet$ & 0.286 $\pm$ 0.007 $\bullet$ & 0.220 $\pm$ 0.014 $\bullet$ & 0.348 $\pm$ 0.006 $\bullet$ & 0.386 $\pm$ 0.005 $\circ$ & 0.697 $\pm$ 0.015 $\circ$ & 0.721 $\pm$ 0.023 $\bullet$ & 0.382 $\pm$ 0.013 $\bullet$ & 0.294 $\pm$ 0.020 $\bullet$ & 0.635 $\pm$ 0.005 $\circ$ & 0.695 $\pm$ 0.031 $\bullet$

         \\\hline
         
         HOMI & 0.165 $\pm$ 0.001 & 0.099 $\pm$ 0.035  & 0.276 $\pm$ 0.052 & 0.183 $\pm$ 0.035 & 0.132 $\pm$ 0.000 & 0.571 $\pm$ 0.002 & 0.873 $\pm$ 0.003& 0.699 $\pm$ 0.022 & 0.242 $\pm$ 0.016 & 0.284 $\pm$ 0.026 & 0.691 $\pm$ 0.038 & 0.651 $\pm$ 0.009 \\\hline\hline
    
    \end{tabular}
 
    \label{tab:one}
\end{table*}

\begin{table*}[!htb]\tiny
\setlength{\tabcolsep}{0.85mm}
\renewcommand\arraystretch{1.2} 
    \centering
        \caption{The comparison of different methods on all the data sets with respect to Marco-averaging AUC. $\bullet$ / $\circ$ indicates whether HOMI is superior or same / inferior to the compared algorithm.}
    \begin{tabular}{c l l l l l l l l l l l l}\hline\hline
          \multicolumn{13}{c}{Macro-averaging AUC $\uparrow$}\\\hline
         Approach &\multicolumn{1}{c}{mediamill} &
        \multicolumn{1}{c}{CAL500} & \multicolumn{1}{c}{emotions} & \multicolumn{1}{c}{enron} &
         \multicolumn{1}{c}{bibtex}&\multicolumn{1}{c}{delicious}&\multicolumn{1}{c}{language log}&
         \multicolumn{1}{c}{birds} & \multicolumn{1}{c}{yeast} & \multicolumn{1}{c}{scene} & \multicolumn{1}{c}{corel5k}&
         \multicolumn{1}{c}{corel16k1}\\\hline
        
         ECC &0.776 $\pm$ 0.004 $\bullet$ & 0.497 $\pm$ 0.013 $\bullet$ & 0.815 $\pm$ 0.013 $\circ$ & 0.650 $\pm$ 0.019 $\bullet$ & 0.870 $\pm$ 0.004 $\bullet$ &0.702 $\pm$ 0.002 $\bullet$&0.544 $\pm$ 0.052 $\bullet$&0.732 $\pm$ 0.050 $\bullet$ & 0.650 $\pm$ 0.017 $\bullet$ & 0.878 $\pm$ 0.004 $\bullet$ & 0.531 $\pm$ 0.018 $\bullet$ & 0.658 $\pm$ 0.005 $\bullet$\\

         BR &0.839 $\pm$ 0.004 $\bullet$ & 0.502 $\pm$ 0.007 $\bullet$ & 0.826 $\pm$ 0.023 $\circ$ & 0.619 $\pm$ 0.028 $\bullet$ &0.874 $\pm$ 0.003 $\bullet$ &0.738 $\pm$ 0.003 $\bullet$&0.553 $\pm$ 0.020 $\bullet$ & 0.718 $\pm$ 0.051 $\bullet$ & 0.629 $\pm$ 0.013 $\bullet$ & 0.886 $\pm$ 0.011 $\bullet$ & 0.524 $\pm$ 0.015 $\bullet$ & 0.673 $\pm$ 0.003 $\bullet$\\
         ML-KNN &0.933 $\pm$ 0.000 $\bullet$ & 0.691 $\pm$ 0.020 $\bullet$ & 0.802 $\pm$ 0.010 $\circ$ & 0.836 $\pm$ 0.032 $\bullet$ &0.938 $\pm$ 0.002 $\bullet$ &0.918 $\pm$ 0.002 $\bullet$&0.703 $\pm$ 0.027 $\bullet$& 0.847 $\pm$ 0.004 $\bullet$ & 0.709 $\pm$ 0.008 $\bullet$ & 0.938 $\pm$ 0.007 $\circ$ & 0.790 $\pm$ 0.023 $\bullet$ & 0.922 $\pm$ 0.001 $\bullet$\\
         WRAP &0.851 $\pm$ 0.003 $\bullet$ & 0.555 $\pm$ 0.011 $\bullet$ & 0.835 $\pm$ 0.029 $\circ$& 0.635 $\pm$ 0.014 $\bullet$ &0.870 $\pm$ 0.004 $\bullet$ &0.755 $\pm$ 0.004 $\bullet$& 0.540 $\pm$ 0.040 $\bullet$& 0.815 $\pm$ 0.041 $\bullet$ & 0.688 $\pm$ 0.008 $\bullet$ & 0.906 $\pm$ 0.005 $\circ$ & 0.581 $\pm$ 0.010 $\bullet$ &0.744 $\pm$ 0.008 $\bullet$\\
         MLSF &0.842 $\pm$ 0.003 $\bullet$& 0.528 $\pm$ 0.012 $\bullet$ & 0.816 $\pm$ 0.020 $\circ$ & 0.503 $\pm$ 0.024 $\bullet$ &0.873 $\pm$ 0.006 $\bullet$ &0.730 $\pm$ 0.004 $\bullet$&0.558 $\pm$ 0.036 $\bullet$& 0.660 $\pm$ 0.038 $\bullet$ & 0.631 $\pm$ 0.008 $\bullet$ & 0.892 $\pm$ 0.010 $\circ$ & 0.520 $\pm$ 0.004 $\bullet$&0.669 $\pm$ 0.007 $\bullet$\\
         LFLC &0.832 $\pm$ 0.007 $\bullet$ &0.553 $\pm$ 0.010 $\bullet$ & 0.834 $\pm$ 0.013 $\circ$ & 0.686 $\pm$ 0.028 $\bullet$ & 0.875 $\pm$ 0.005 $\bullet$ &0.757 $\pm$ 0.003 $\bullet$&0.557 $\pm$ 0.004 $\bullet$& 0.835 $\pm$ 0.040 $\bullet$ & 0.684 $\pm$ 0.009 $\bullet$ & 0.924 $\pm$ 0.005 $\circ$ & 0.568 $\pm$ 0.012 $\bullet$ &0.726 $\pm$ 0.004 $\bullet$ \\ 
         BILAS & 0.889 $\pm$ 0.006 $\bullet $& 0.533 $\pm$ 0.000 $\bullet$ &0.758 $\pm$ 0.000 $\bullet$ & 0.844 $\pm$ 0.000 $\bullet$ &0.920 $\pm$ 0.004 $\bullet$ &0.796 $\pm$ 0.005 $\bullet$& 0.559 $\pm$ 0.006 $\bullet$ & 0.495 $\pm$ 0.000 $\bullet$ &0.650 $\pm$ 0.000 $\bullet$ &0.920 $\pm$ 0.000 $\circ$ &0.655 $\pm$ 0.037 $\bullet$ &0.686 $\pm$ 0.024 $\bullet$  \\ 
         GLOCAL & 0.605 $\pm$ 0.009 $\bullet$ & 0.509 $\pm$ 0.030 $\bullet$ & 0.750 $\pm$ 0.032 $\bullet$ & 0.787 $\pm$ 0.022 $\bullet$ & 0.825 $\pm$ 0.021 $\bullet$ &0.637 $\pm$ 0.004 $\bullet$&0.613 $\pm$ 0.023 $\bullet$& 0.513 $\pm$ 0.027 $\bullet$ & 0.732 $\pm$ 0.040 $\bullet$ & 0.826 $\pm$ 0.004 $\bullet$ & 0.518 $\pm$ 0.043 $\bullet$ & 0.581 $\pm$ 0.032 $\bullet$ \\

         ML-LRC & 0.957 $\pm$ 0.003 $\bullet$ & 0.849 $\pm$ 0.003 $\bullet$ & 0.726 $\pm$ 0.006 $\bullet$ & 0.949 $\pm$ 0.013 $\bullet$ & 0.989 $\pm$ 0.000 $\bullet$ &0.980 $\pm$ 0.001 $\bullet$&0.927 $\pm$ 0.004 $\bullet$ & 0.944 $\pm$ 0.006 $\bullet$ & 0.704 $\pm$ 0.000 $\bullet$ & 0.861 $\pm$ 0.002 $\bullet$ & 0.990 $\pm$ 0.000 $\bullet$ & 0.981 $\pm$ 0.000 $\bullet$\\
         
         CLML & 0.950 $\pm$ 0.008 $\bullet$ & 0.534 $\pm$ 0.010 $\bullet$ & 0.684 $\pm$ 0.019 $\bullet$ & 0.664 $\pm$ 0.031 $\bullet$ & 0.990 $\pm$ 0.000 $\bullet$ & 0.980 $\pm$ 0.000 $\bullet$ & 0.905 $\pm$ 0.041 $\bullet$ & 0.944 $\pm$ 0.008 $\bullet$ & 0.695 $\pm$ 0.004 $\bullet$ & 0.820 $\pm$ 0.003 $\bullet$ & 0.892 $\pm$ 0.014 $\bullet$ & 0.981 $\pm$ 0.000 $\bullet$
         
         \\\hline
         
         HOMI & 0.969 $\pm$ 0.000 & 0.865 $\pm$ 0.006  & 0.795 $\pm$ 0.023 & 0.950 $\pm$ 0.012 & 0.990 $\pm$ 0.000 & 0.981 $\pm$ 0.002 & 0.985 $\pm$ 0.001& 0.948 $\pm$ 0.006 & 0.800 $\pm$ 0.004 & 0.887 $\pm$ 0.005 & 0.991 $\pm$ 0.000 & 0.981 $\pm$ 0.000\\\hline\hline
    
    \end{tabular}

    \label{tab:auc}
\end{table*}

\subsection{Data Sets}
In this section, comparative studies were conducted on twelve commonly used benchmark multi-label data sets.
Table \ref{tab:chara} summarizes the detailed characteristics of each data set $\mathcal{D}$, with the number of examples ($n$), the dimension of features ($m$), the number of class labels ($l$), label cardinality, i.e., average number of relevant labels per example ($LCard(\mathcal{D})$), label density, i.e. label cardinality over the number of class labels ($LDen(\mathcal{D})$), and number of distinct label sets ($D L(\mathcal{D})$) in $\mathcal{D}$. Those data sets are publicly available at \url{https://mulan.sourceforge.net/datasets-mlc.html}

\subsection{Compared Methods}
We compared HOMI with the following ten state-of-the-art multi-label classification approaches.

\begin{itemize}
    \item ECC (Ensemble of classifier chains) \cite{2009Classifier1212} is an ensemble-based multi-label classification approach, building an ensemble of N classifier chains to solve multi-label classification. [hyper-parameter configuration: N = 5];

    \item BR \cite{BOUTELL2004Learning} is a classical algorithm in multi-label classification, trying to decompose the original multi-label classification task into a set of binary classification tasks. [hyper-parameter C = 1];
    \item ML-KNN \cite{zhang2007mlds} is a popular first-order multi-label learning algorithm based on k-nearest-neighbour classification. [hyper-parameter configuration: k=10];
    \item WRAP \cite{yu2021multiwrgweg} tires to generate label specific features in an embedded feature space to deal with multi-label classification. [hyper-parameter configuration: step\_size = 1, $\lambda$ = 0.1, $\alpha$ = 0.9, $d = \alpha \min (m,l)$];
    \item MLSF \cite{sun2016multiasdf} generates label-specific features by analyzing local and global feature-to-label correlations. [hyper-parameter configuration: K = $l$/10, $\epsilon$ = 0.01, $\alpha$ = 0.8, $\gamma$ = 0.01];
    \item LFLC \cite{ma2019multilabelfgsdgqee} generates label tailored features by analyzing local
    and global feature-to-label correlations. [hyper-parameter configuration: grid search for $\lambda \in \{1, 3,..., 19\} $ with step-size 2, $\eta \in \{1e-10,..., 1e-5\} $ with a multiple of $e$ at each step, $\beta = 10^4$];
    \item BILAS \cite{zhang2021bilabel} generates a group of tailored features for a pair of class labels with heuristic prototype selection and embedding. [hyper-parameter configuration: t=0.1, ratio=0.5];
    \item GLOCAL \cite{2017Multisdfsaf} uses low-rank factorization to dig the global and local label correlations at the same time, through learning a latent label representation and optimizing label manifolds. [hyper-parameter configuration: $\lambda = 1$];
    
    \item ML-LRC \cite{9084698} is a low-rank approach applying low-rank constraints to the label matrix to mine the local correlations of class labels. [hyper-parameter configuration: grid search for $\alpha, \beta \in \{2^{-10}, 2^{-9}, 2^{-8},...,2^{10}\} $, $\gamma = 0.1$ and $\tau=0.5$];

    \item CLML \cite{li2022learning} is a approach that learns common and label-specific features based on the correlation information from labels and instances. [hyper-parameter configuration: grid search for $\alpha, \beta, \lambda_1, \lambda_2 \in \{2^{-10}, 2^{-9}, 2^{-8},...,2^{10}\}$ with step $2^2$].
\end{itemize}

In brief, BR \cite{BOUTELL2004Learning} and ML-KNN \cite{zhang2007mlds} belong to first-order approaches, BILAS \cite{zhang2021bilabel} is a second-order approach, and WRAP \cite{yu2021multiwrgweg}, ECC \cite{2009Classifier1212}, GLOCAL \cite{2017Multisdfsaf} and CLML \cite{li2022learning} are high-order approaches. \textit{Particularly, GLOCAL and ML-LRC \cite{9084698} are two low-rank based approaches.} MLSF \cite{sun2016multiasdf} and LFLC  \cite{ma2019multilabelfgsdgqee} are two label-specific approaches. The hyper-parameter configurations for different methods are suggested by their original papers. 

\subsection{Experimental Settings}

The hyper parameters of HOMI are set as follows: $\beta=2$, $\gamma=1$, $\lambda=1$, $iter=100$ and $s=10$\footnote{For emotions and bibtex, $s$ is set to be 2.}. Following \cite{yu2021multiwrgweg}, five-fold cross-validation is performed on each data set, with mean metric and standard deviation recorded.

\subsection{Evaluation Metrics}
Let $C_i^+, C_i^-$ be the sets of positive and negative labels corresponding to the $i$th instance, and $T_i^+, T_i^-$ be the sets of positive and negative instances corresponding to the $i$th label, $p$ is the number of the test instances. We chose the following four popular metrics to evaluate the performance of the proposed method and the methods under comparison.
\begin{itemize}
    \item Hamming loss (Hloss) evaluates the rate of the mistook labels. Hloss = $\frac{1}{p}\sum_i f(\mathbf{x}_i)\Delta \mathbf{y_i}$, $\Delta$ stands for the symmetric difference between two sets, i.e., $a\Delta b=1$ (resp. 0) if $a=b$ (resp. a $\neq$ b).
    \item Ranking loss (Rloss) calculates the fraction that a negative label is ranked higher than a positive label. Specifically, for instance $i$, suppose $M_i = \{ ( j^{'}, j^{''} ) |\mathbf{g}_{j^{'}}(\mathbf{x}_i) \le \mathbf{g}_{j^{''}}(\mathbf{x}_i), (j^{'}, j^{''})\in C_i^+\times C_i^- \}$, Rloss = $\frac{1}{p} \sum_{i=1}^p \frac{|M_i|}{|C_i^+\|C_i^-|}$.
    \item One-error evaluates the fraction of examples whose top-ranked label is not in the relevant label set. One-error = $\frac{1}{p}\sum_{i=1}^p[c_{argmax_j f_j(\mathbf{x}_i)} \notin \mathbf{y_i} ]$.
    \item  Average Area Under the ROC Curve (Macro-averaging AUC) denotes the fraction that a positive instance is ranked higher than a negative instance averaged over all labels. Suppose $N_i = \{(i^{'}, i^{''}) | \mathbf{g}_{j}(x_{i^{'}}) \ge \mathbf{g}_j(x_{i^{''}}), (x_{i^{'}}, x_{j^{''}} \in T_j^+ \times T_j^-) \}$, Macro-averaging AUC = $\frac{1}{l} \sum_{j=1}^l \frac{|N_i|}{|T_i^+\|T_i^-|}$.
\end{itemize}

For hamming loss, ranking loss and one-error, the lower the better, while for macro-averaging AUC, the higher the better. All the metrics lie in the range of $[0,1]$.

\subsection{Experimental Analysis}
Tables \ref{tab:hamming}-\ref{tab:auc} show the experimental results of the proposed method and the compared baselines on twelve data sets with respect to four metrics. Additionally, the widely-used \textit{Friedman test}\cite{2006Statistical6236} is used for statistical analysis of the performance among all the methods on the benchmark data sets. Suppose $k$ denotes the number of comparing algorithms, $N$ denotes the number of data sets and $r_i^j$ denotes the rank of the $j$th approach on the $i$th data set. Suppose $R_j = \frac{1}{N}\sum_{i=1}^Nr_i^j$ denotes the average rank of the $j$th method on all the data sets. The Friedman statistic $F_F$, which is distributed according to the $F$-distribution with $(k-1)$ numerator degrees of freedom and $(k-1)(N-1)$ denominator degrees of freedom, is defined as

$$
F_F = \frac{(N-1)\mathcal{X}_F^2}{N(k-1) - \mathcal{X}_F^2},$$where

$$
\mathcal{X}_F^2 = \frac{12N}{k(k+1)} ( \sum^k_{j=1}R_j^2 - \frac{k(k+1)^2}{4} ).
$$

Table \ref{tab:friedman} reports the detailed statistics over all evaluation metrics as well as the related critical value at 0.05 significance level for HOMI ($k=11, N=12$). We can observe that the $F_F$ value is larger than the critical value w.r.t. all evaluation metrics. Therefore, the null hypothesis of equal performance among comparing approaches is clearly rejected.

\begin{table}[ht]
\setlength{\tabcolsep}{1.2mm}\renewcommand\arraystretch{1.2} 
    \centering

    \begin{tabular}{c c c}
     \hline \hline
    
       Evaluation Metric &  $F_F$ & Critical Value\\\hline
       Hamming Loss & 6.0589 &\multirow{4}{*}{1.9178}\\
       Ranking Loss & 6.8014 \\
       One-error & 7.2527 \\
       Macro-averaging AUC & 10.1662\\
        \hline\hline
    \end{tabular}
    \caption{Friedman test statistics over each evaluation metrics and the critical value at 0.05 significance level ($k=11,N=12$).}
    \label{tab:friedman}
\end{table}

In order to verify whether HOMI significantly outperforms other algorithms, we employ \textit{Holm's procedure} \cite{2006Statistical6236} as the post-hoc test by treating HOMI as the control approach. Without loss of generality, we take HOMI as the first comparing approach $\mathcal{A}_1$, and for the other $k-1$ approaches, we let $\mathcal{A}_j$ $(2\le j\le k)$ denote the one with the $(j-1)$th largest average rank. Then, the test statistic for comparing $\mathcal{A}_1$ and $\mathcal{A}_j$ is defined as follows:

$$
z_j = (R_1-R_j)/\sqrt{\frac{k(k+1)}{6N}} \quad (2\le j\le k),
$$

Accordingly, let $p_j$ denote the $p$-value of $z_j$ under normal distribution, and the Holm's procedure sequentially checks whether $p_j$ is below $\alpha/(k-j+1)$ in ascending order of $j$ at significance level $\alpha$. Specifically, the Holm's procedure is supposed to terminate at $j^*$ where $j^*$ is the first $j$ that satisfying $p_j \ge \alpha/(k-j+1)$\footnote{If $p_j<\alpha/(k-j+1)$ holds for all $j$, $j^*$ is set to be $k+1$.}. Then HOMI is deemed to perform significantly different compared with $\mathcal{A}_j$ where $j\in\{2,...,j^*-1\}$.

Table \ref{tab:holm} reports the statistics by taking Holm's procedure as post-hoc test at 0.05 significance level, where HOMI is treated as the control approach. We can have the following observations based on the experimental results:

\begin{figure*} [ht!]
	\centering
	\subfigure[\label{fig:alabel}]{
		\includegraphics[scale=0.35]{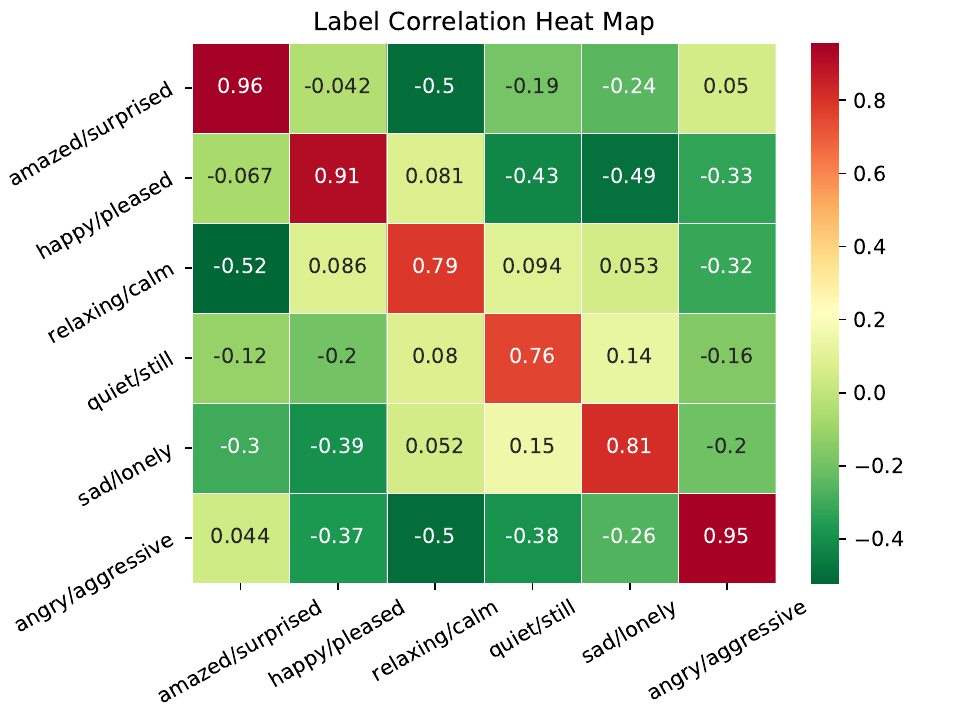}}
    \subfigure[\label{fig:blabel}]
    {
		\includegraphics[scale=0.21]{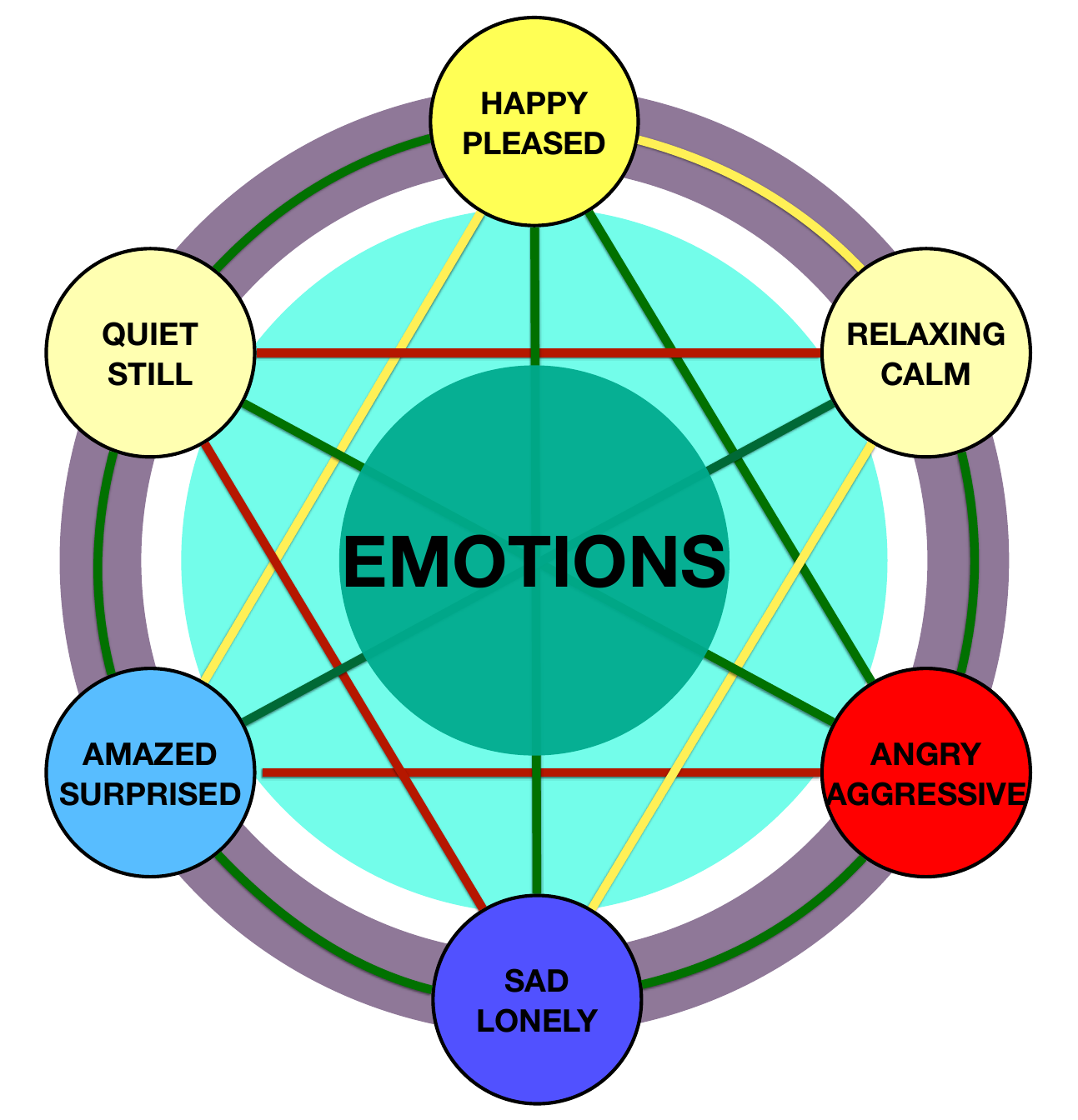}}
	\subfigure[\label{fig:clabel}]{
	\includegraphics[scale=0.17]{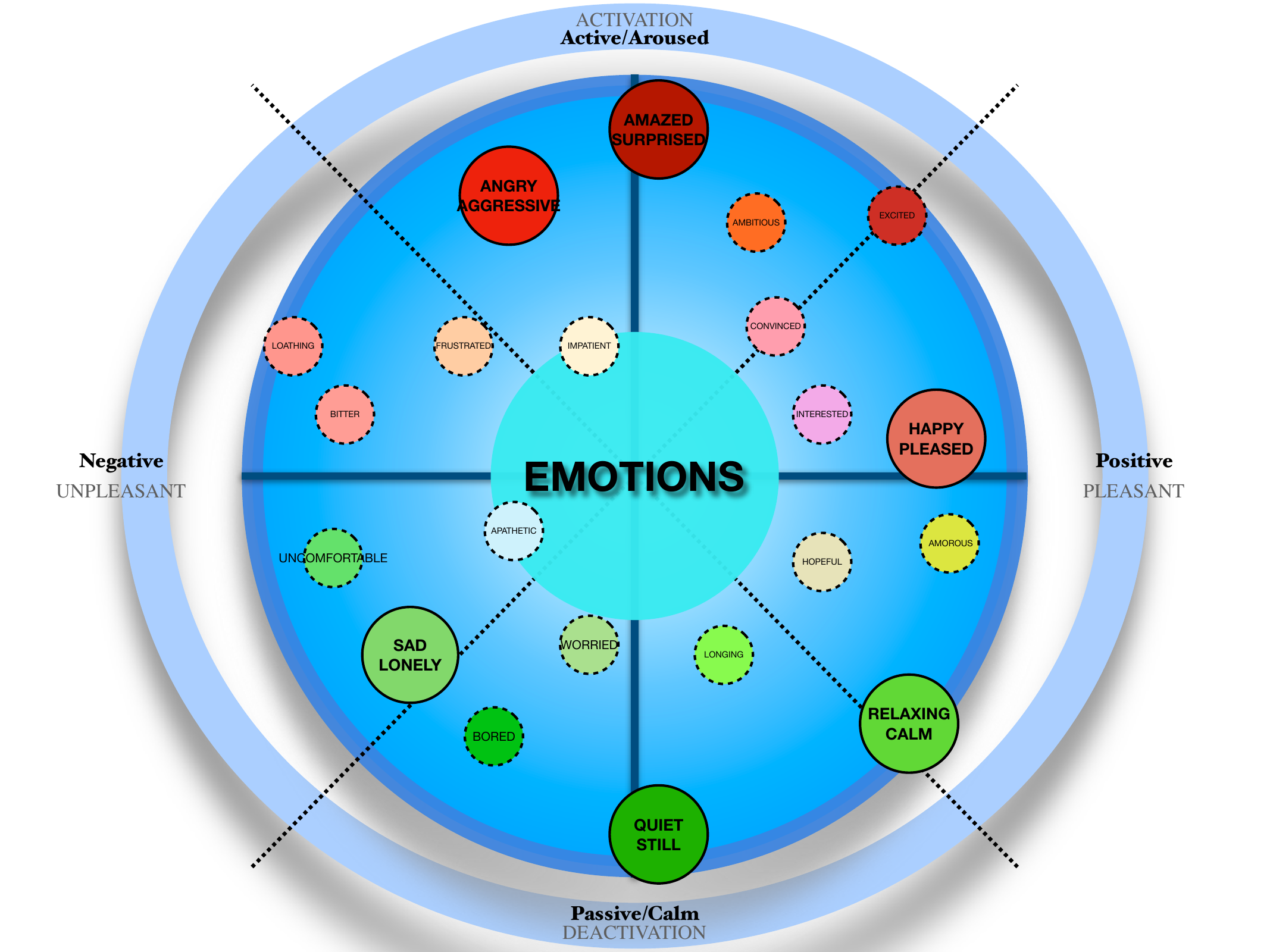}}

	\caption{(a) The normalized matrix $\mathbf{B}$ learned by HOMI on emotions data set. Redder color block represents larger value, while greener color block represents smaller value. (b) Label correlations based on matrix $\mathbf{B}$. The green line between two labels indicates that the two labels are negatively correlated, while red one indicates those are positively related, and yellow one means that the correlation is uncertain or very weak. (c) Russell's emotion circumplex \cite{russell1980circumplex}. The large solid circles represent the labels in data set emotions, and the small dotted circles represent other representative emotions do not exist in the emotions data set.}
	\label{fig:my_label} 
\end{figure*}

\begin{figure*} [ht!]
	\centering
	\subfigure[\label{fig:ah}][Hamming Loss]{
		\includegraphics[scale=0.26]{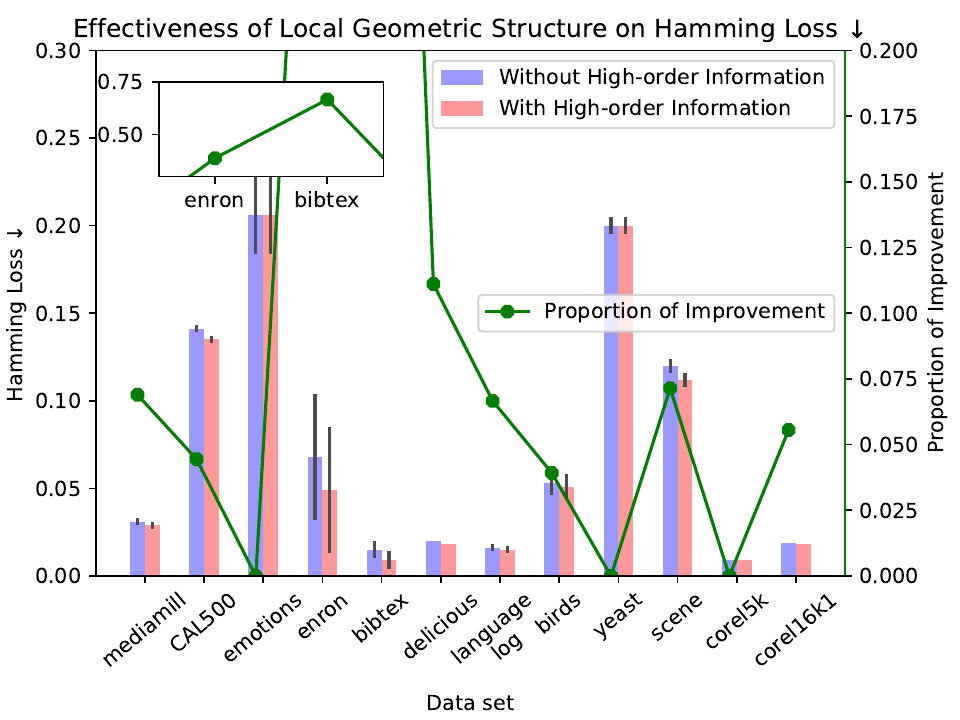}}
	\subfigure[\label{fig:bh}][Ranking Loss]{
		\includegraphics[scale=0.26]{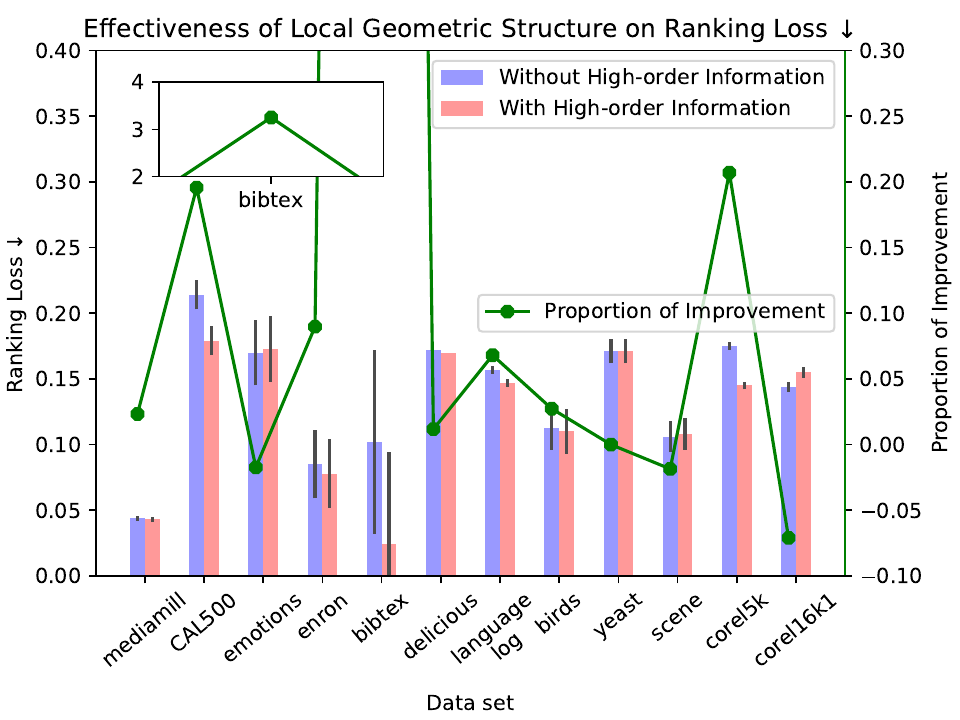}}
	\subfigure[\label{fig:ch}][One-error]{
		\includegraphics[scale=0.26]{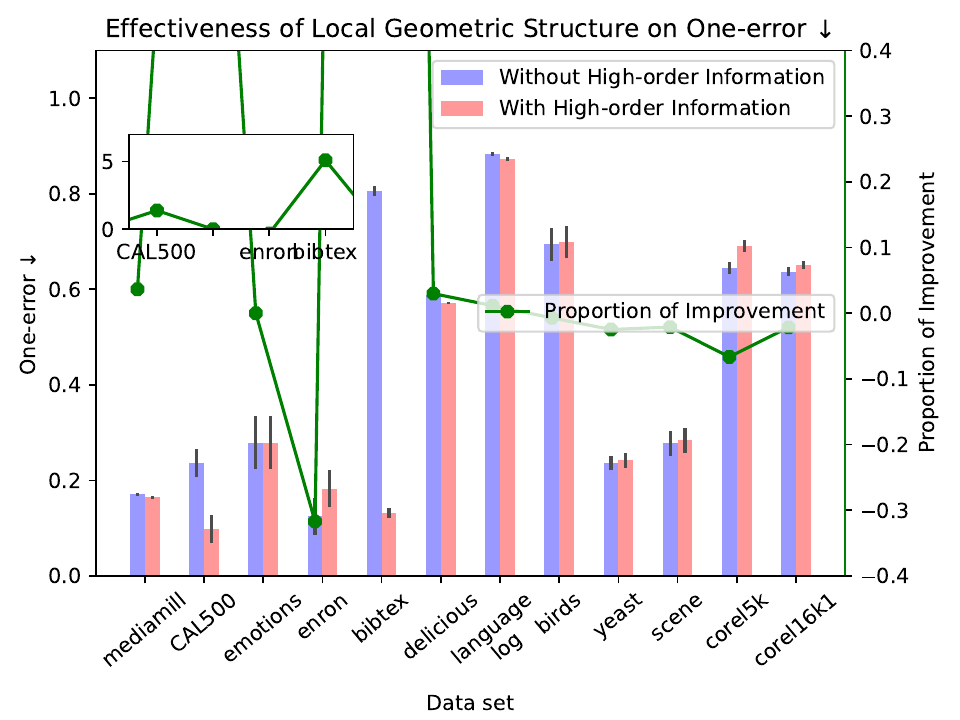}}
	\subfigure[\label{fig:dh}][Macro-averaging AUC]{
		\includegraphics[scale=0.26]{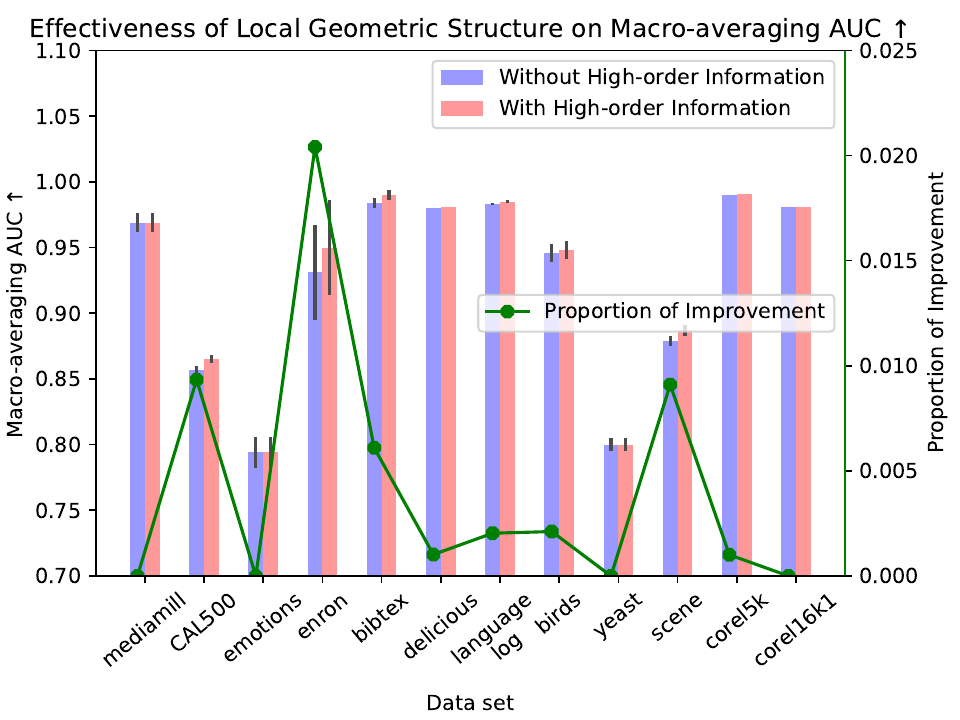} }

	\caption{Usefulness of high-order information. The bar in each sub-figure stands for the value of each metric on different data sets based on the left y-axis. The black vertical line on the top of the bars denotes the standard deviation on each metric. The green line in each sub-figure stands for the proportion of improvement based on the right y-axis.}
	\label{fig high} 
\end{figure*}

\begin{figure*} [ht!]
	\centering
	\subfigure[\label{fig:aj}][Hamming Loss]{
		\includegraphics[scale=0.26]{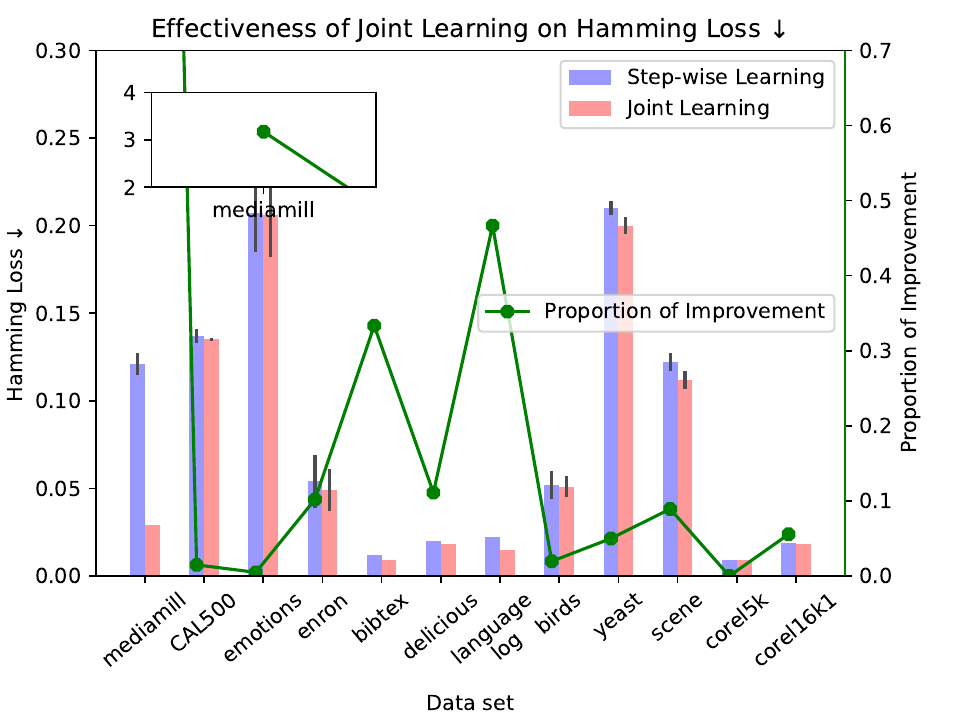}}
	\subfigure[\label{fig:bj}][Ranking Loss]{
		\includegraphics[scale=0.26]{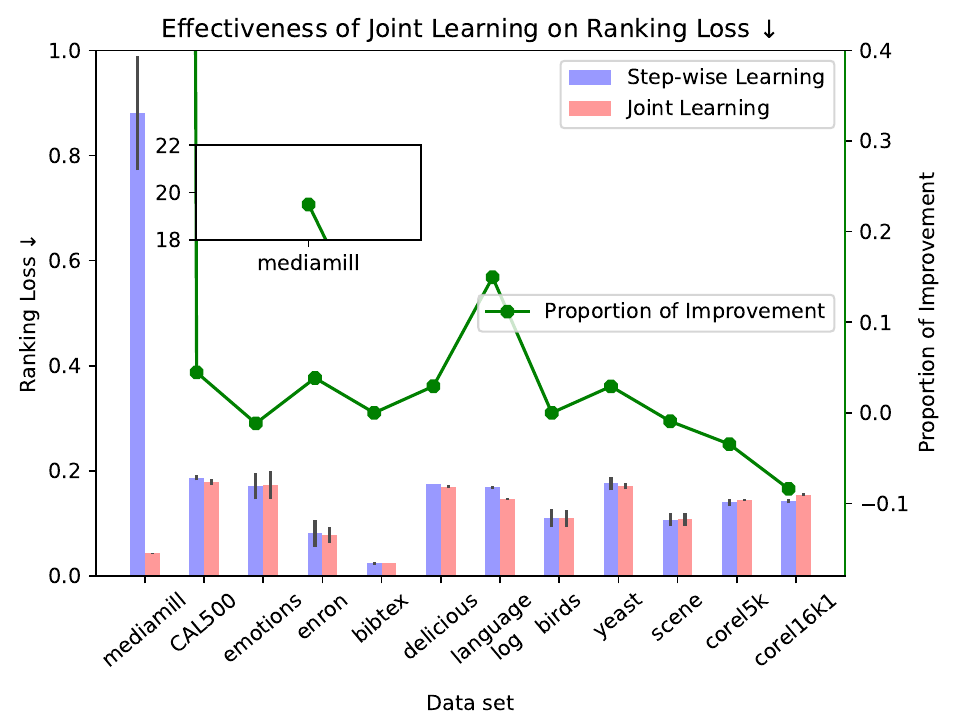}}
	\subfigure[\label{fig:cj}][One-error]{
		\includegraphics[scale=0.26]{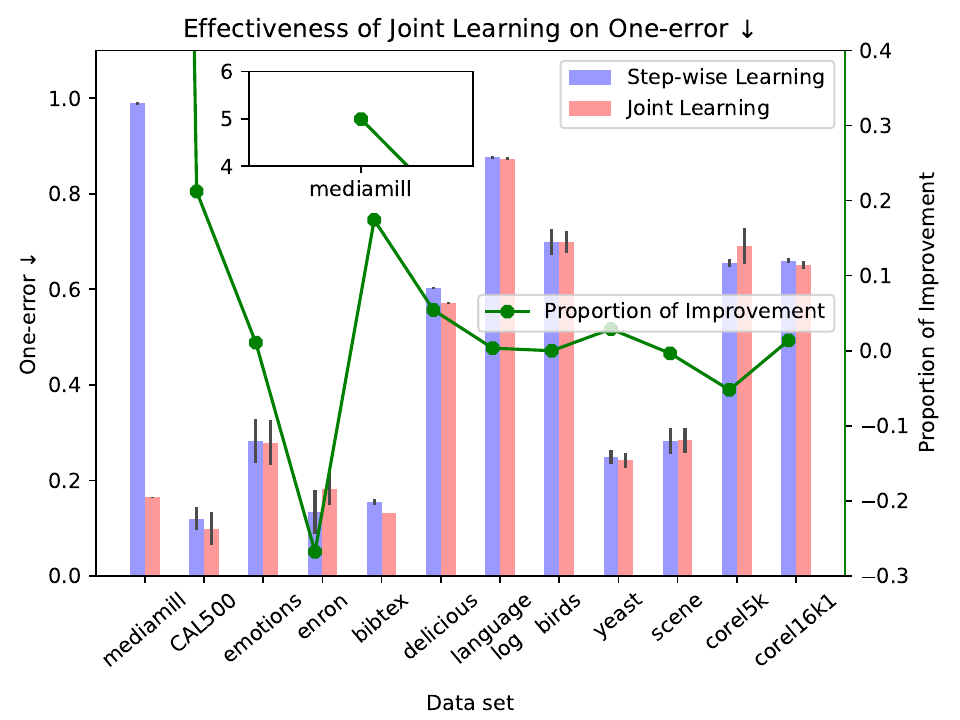}}
	\subfigure[\label{fig:dj}][Macro-averaging AUC]{
		\includegraphics[scale=0.26]{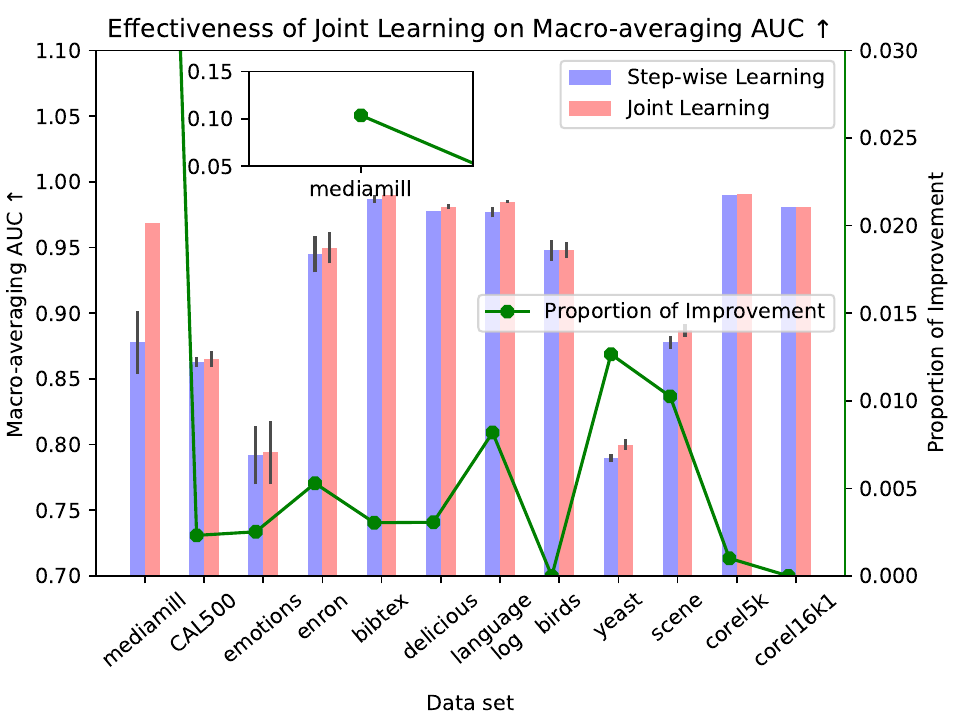} }

	\caption{Effectiveness of joint learning. The bar in each sub-figure stands for the value of each metric on different data sets based on the left y-axis. The black vertical line on the top of the bars denotes the standard deviation on each metric. The green line in each sub-figure stands for the proportion of improvement based on the right y-axis.}
	\label{fig joint} 
\end{figure*}

\begin{figure*} [ht!]
    \centering
	\subfigure[\label{fig:al}][Hamming Loss]{
		\includegraphics[scale=0.26]{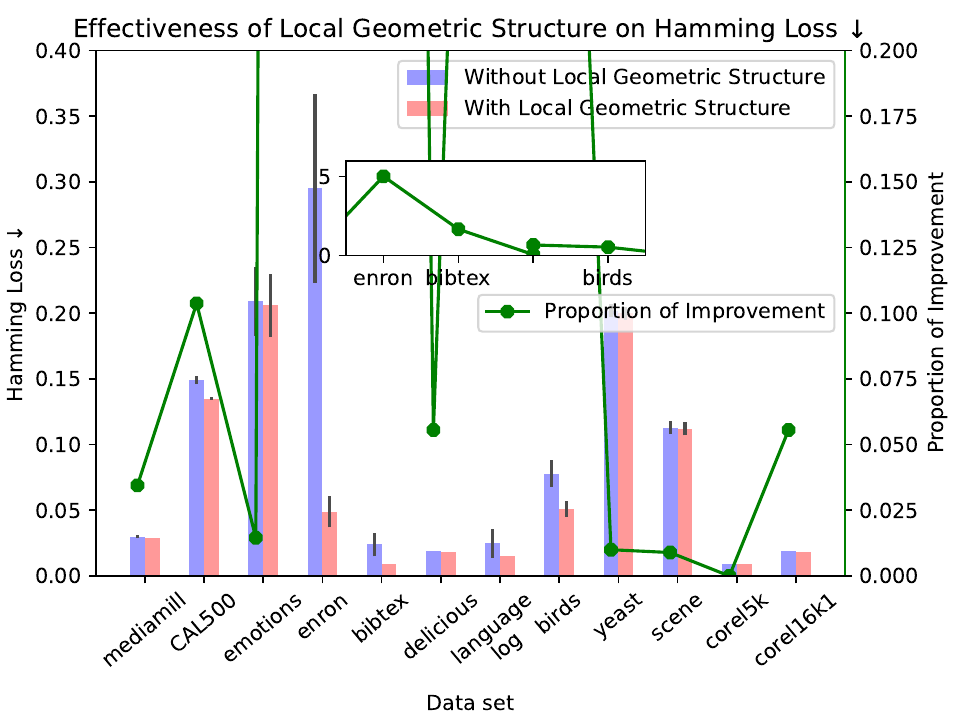}}
	\subfigure[\label{fig:bl}][Ranking Loss]{
		\includegraphics[scale=0.26]{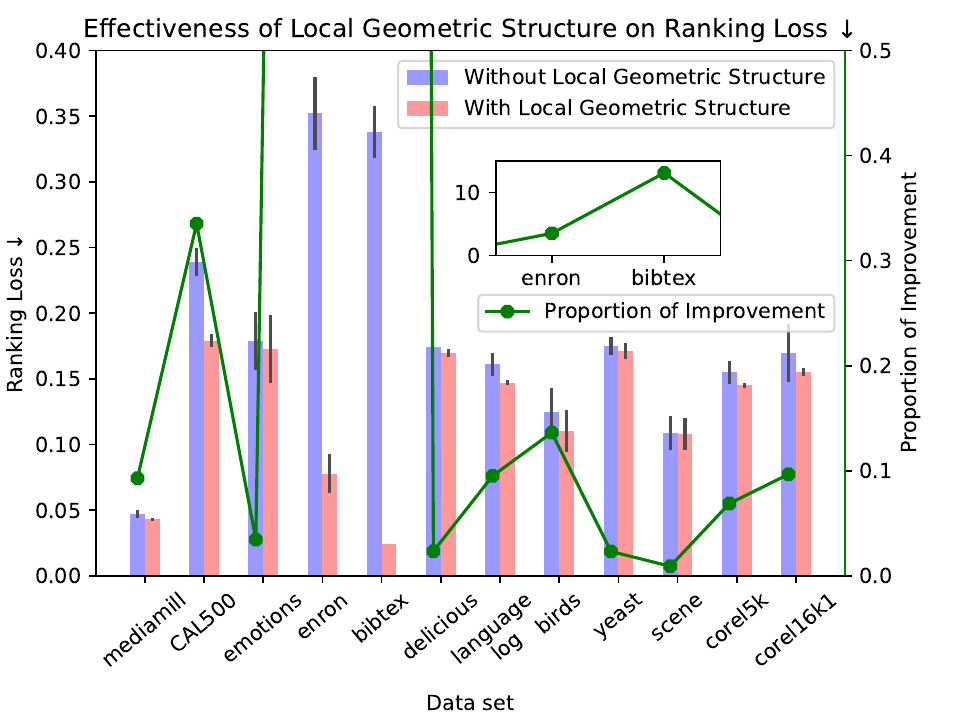}}
	\subfigure[\label{fig:cl}][One-error]{
		\includegraphics[scale=0.26]{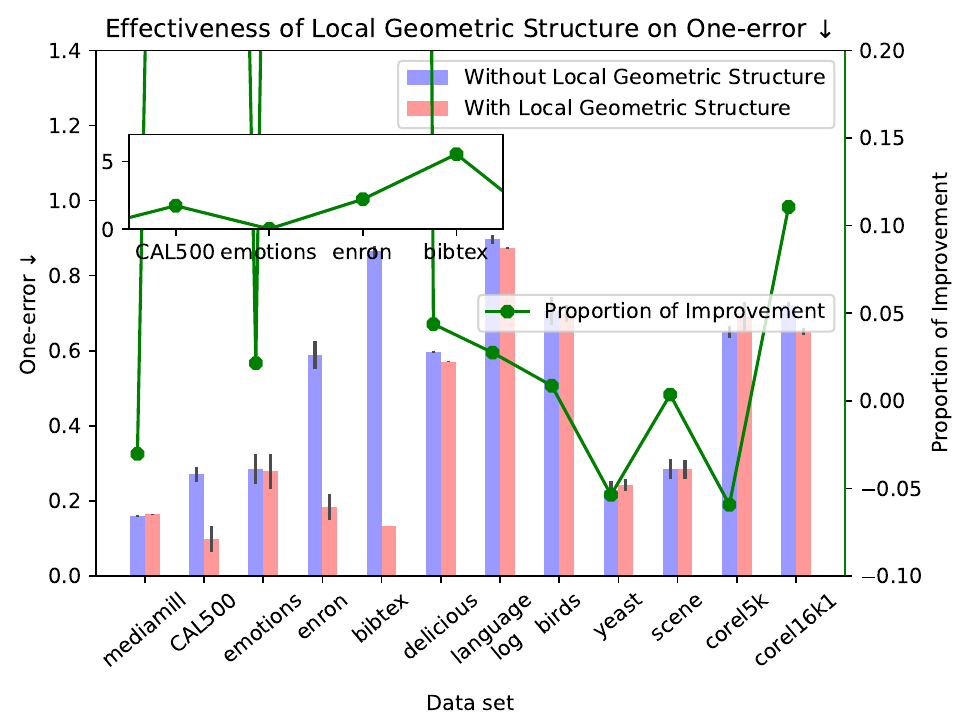}}
	\subfigure[\label{fig:dl}][Macro-averaging AUC]{
		\includegraphics[scale=0.26]{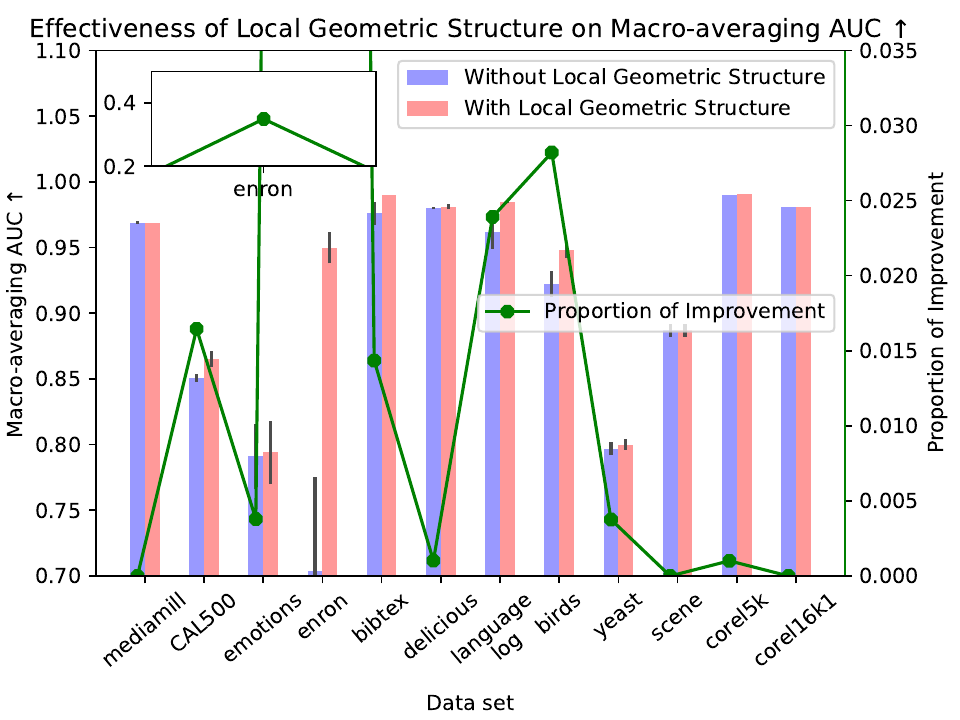}}

	\caption{Usefulness of local geometric structure. The bar in each sub-figure stands for the value of each metric on different data sets based on the left y-axis. The black vertical line on the top of the bars denotes the standard deviation on each metric. The green line in each sub-figure stands for the proportion of improvement based on the right y-axis.}
	\label{fig geometric} 
\end{figure*}

\begin{figure*} [ht!]
	\centering
	\subfigure[\label{fig:af}][$\beta$]{
		\includegraphics[scale=0.26]{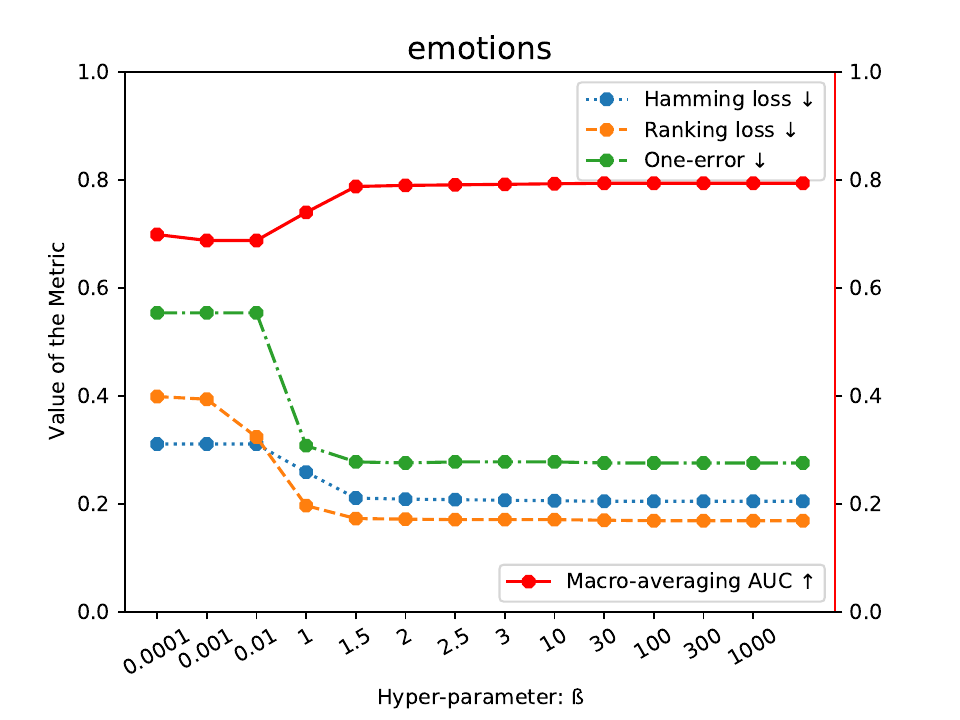}}
	\subfigure[\label{fig:bf}][$\gamma$]{
		\includegraphics[scale=0.26]{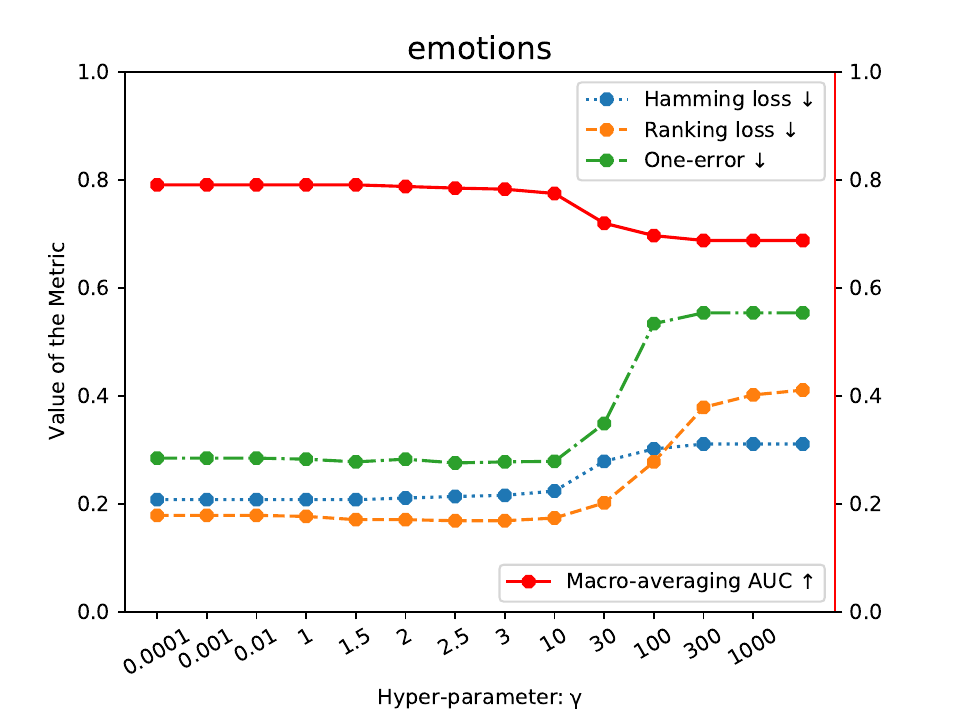}}
	\subfigure[\label{fig:cf}][$\lambda$]{
		\includegraphics[scale=0.26]{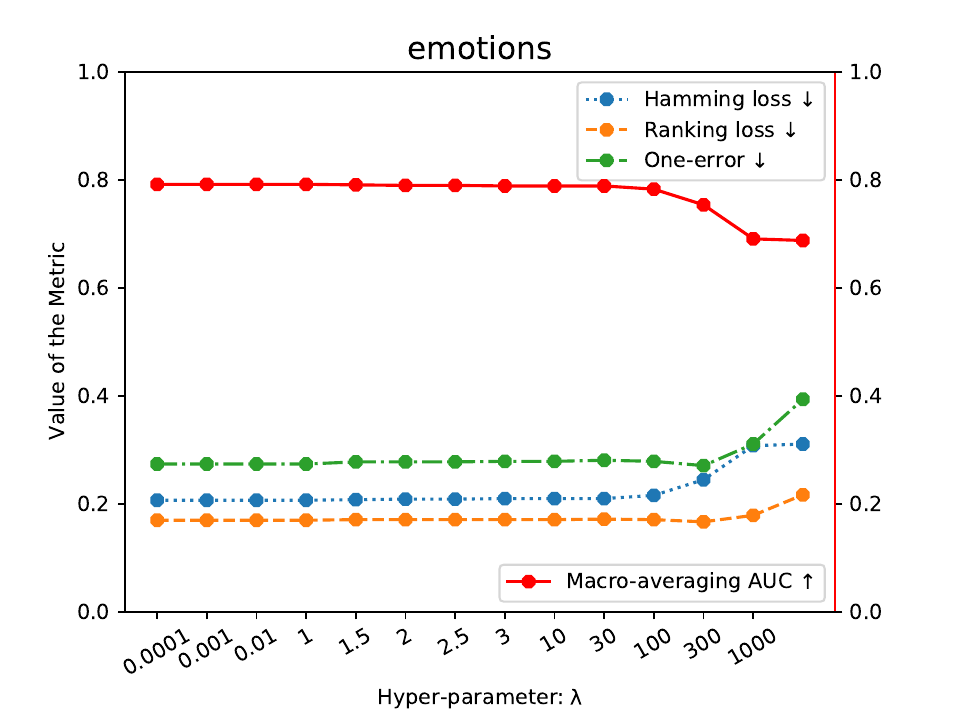}}
	\subfigure[\label{fig:df}][$s$]{
		\includegraphics[scale=0.26]{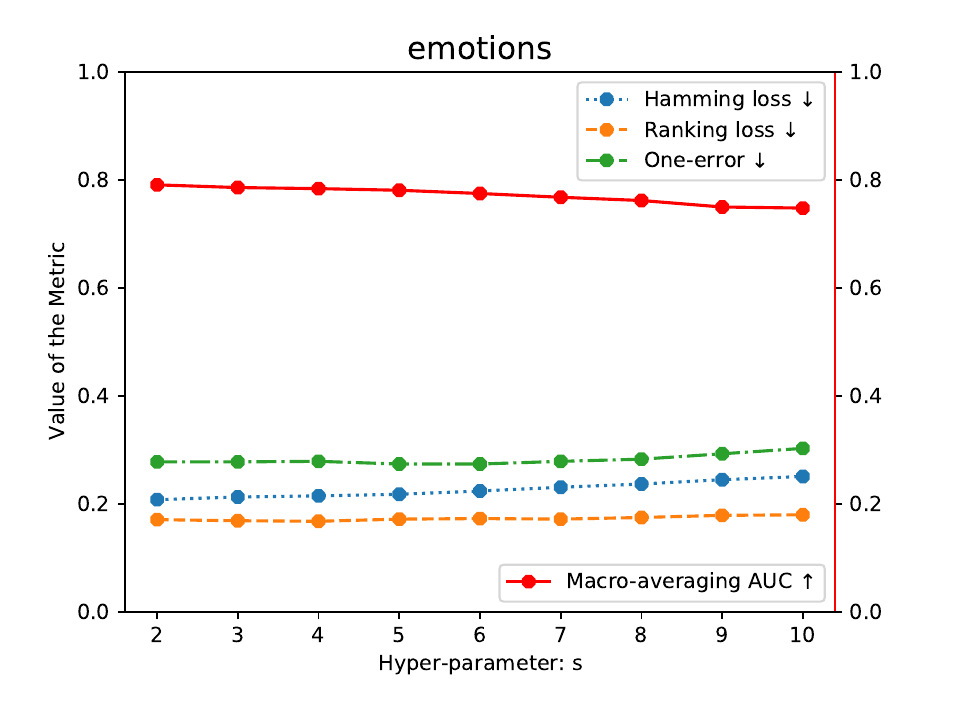} }

	\caption{Sensitivity analysis. The value of macro-averaging AUC is based on the right y-axis, while the other three metrics are based on the left one.}
	\label{fig further} 
\end{figure*}

\begin{figure*} [ht!]
	\centering
	\subfigure[\label{fig:ac}][mediamill]{
		\includegraphics[scale=0.17]{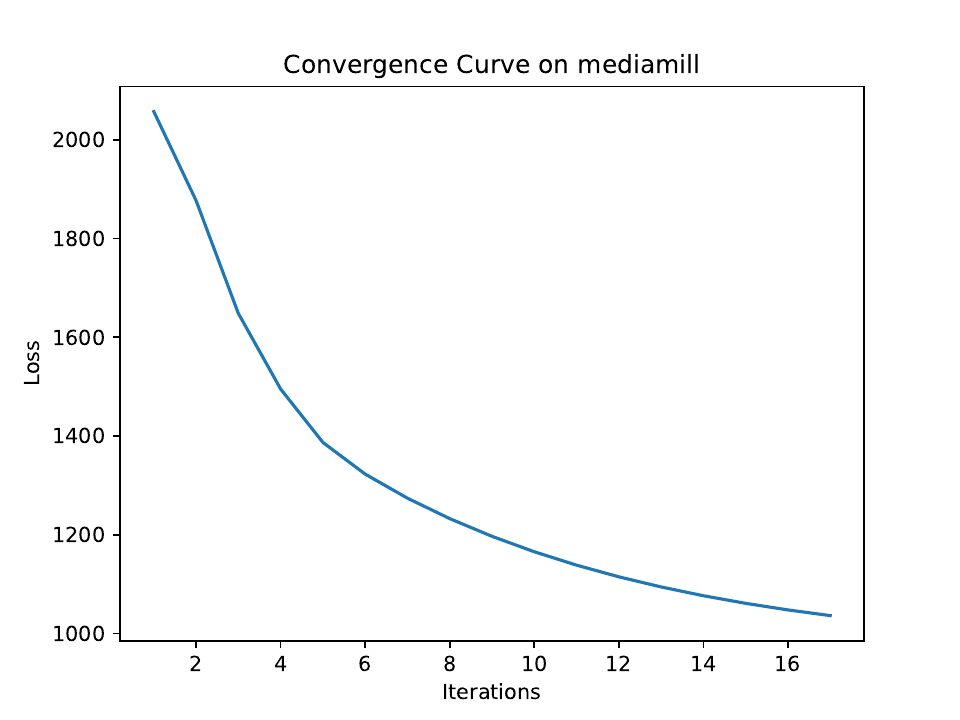}}
	\subfigure[\label{fig:bc}][CAL500]{
		\includegraphics[scale=0.17]{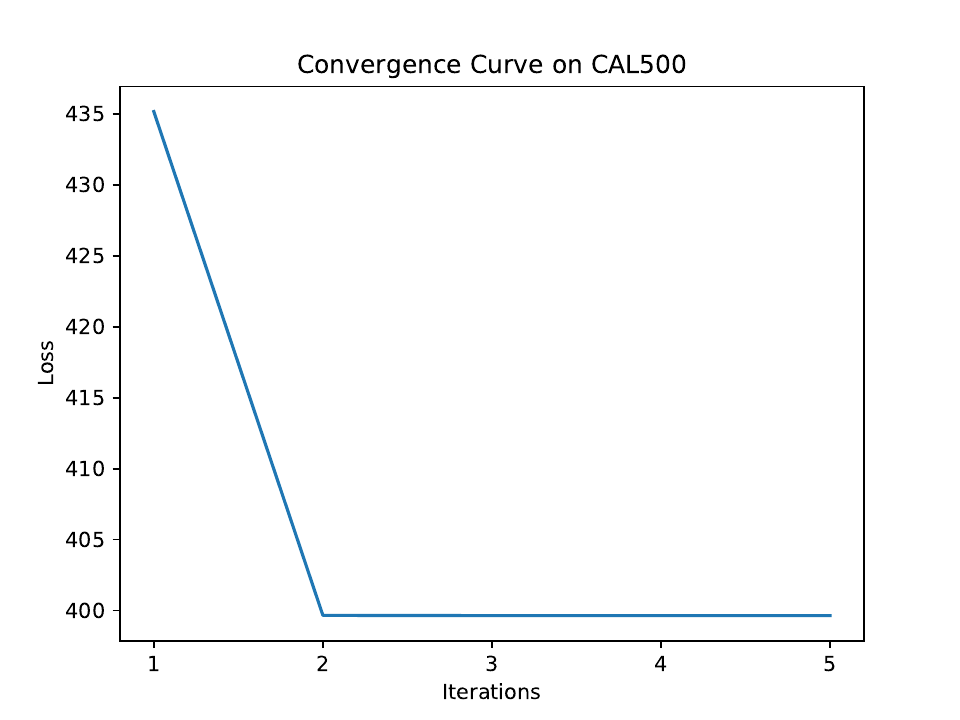}}
	\subfigure[\label{fig:ac}][emotions]{
		\includegraphics[scale=0.17]{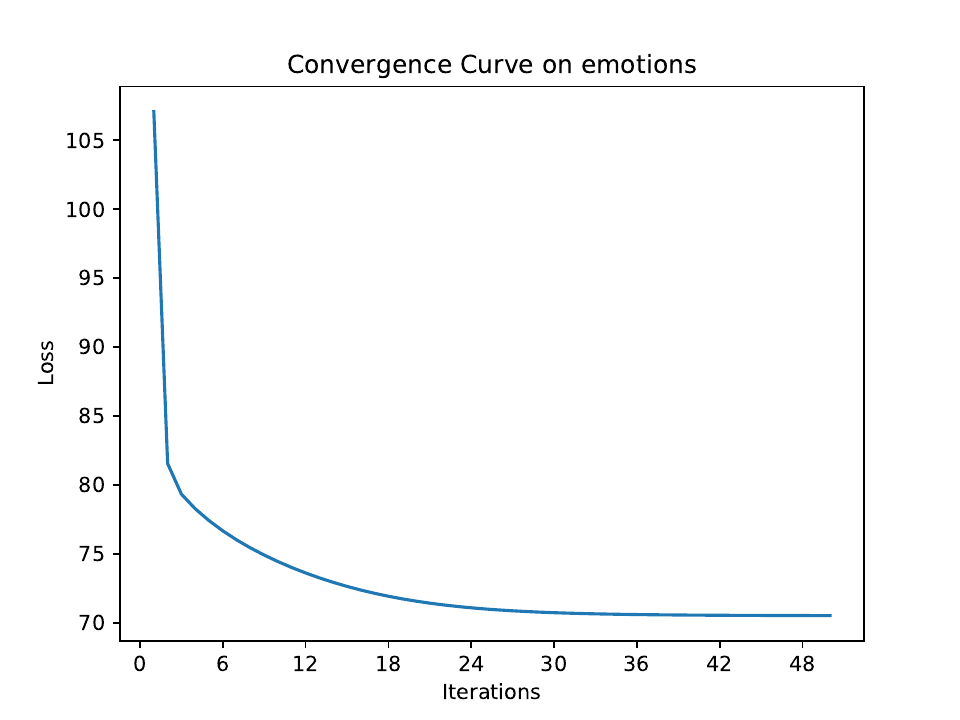}}
	\subfigure[\label{fig:bc}][enron]{
		\includegraphics[scale=0.17]{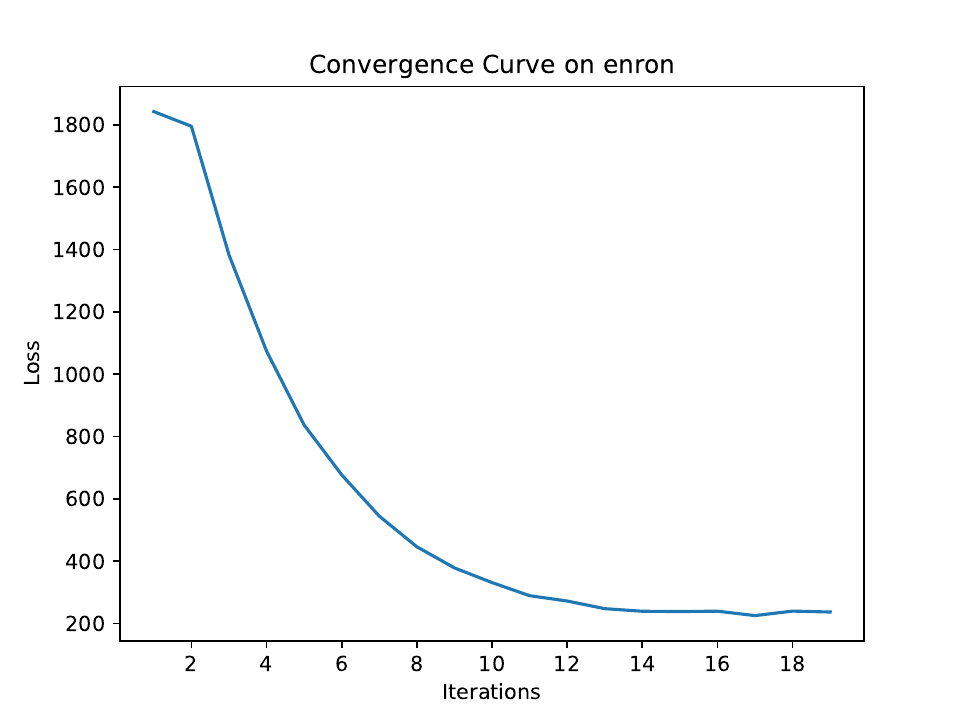}}
	\subfigure[\label{fig:bc}][bibtex]{
		\includegraphics[scale=0.17]{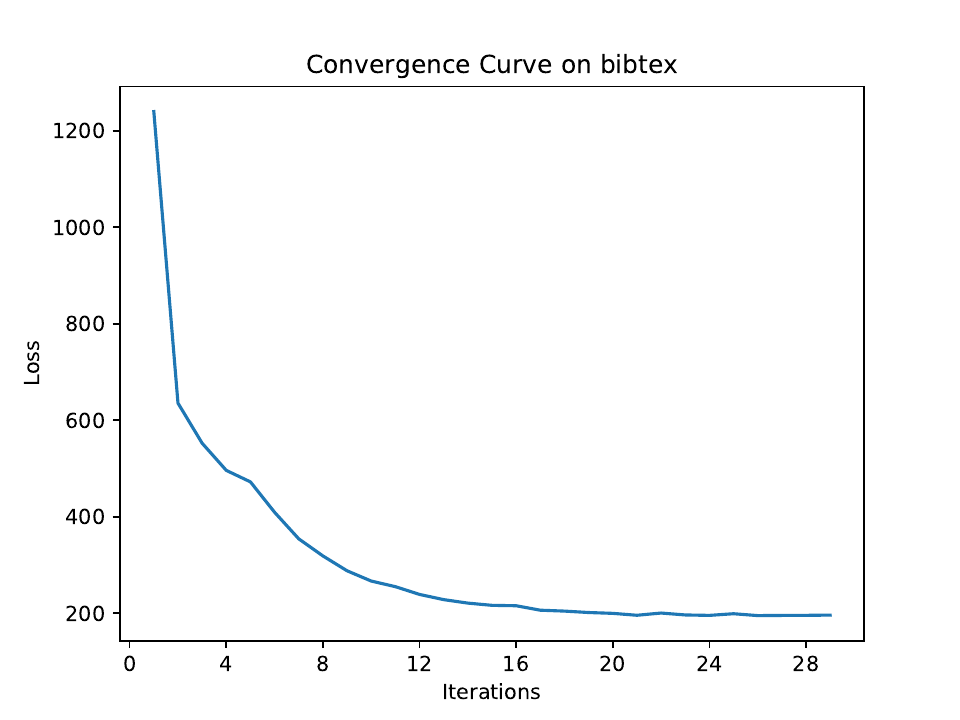}}
    \subfigure[\label{fig:bc}][delicious]{
		\includegraphics[scale=0.17]{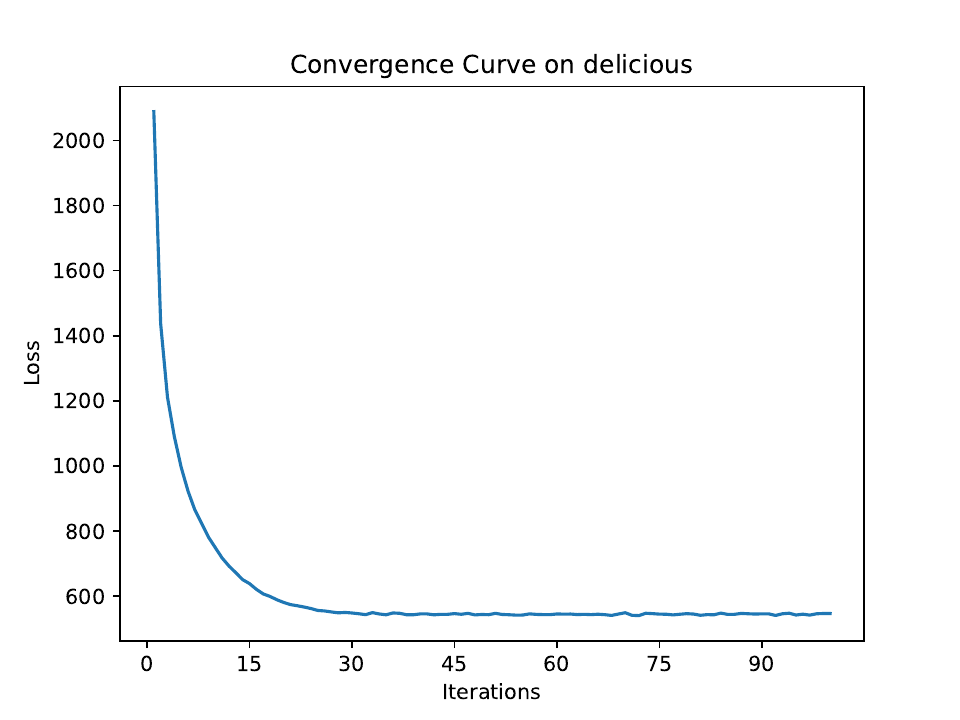}}
	\\
     \subfigure[\label{fig:ac}][language log]{
		\includegraphics[scale=0.17]{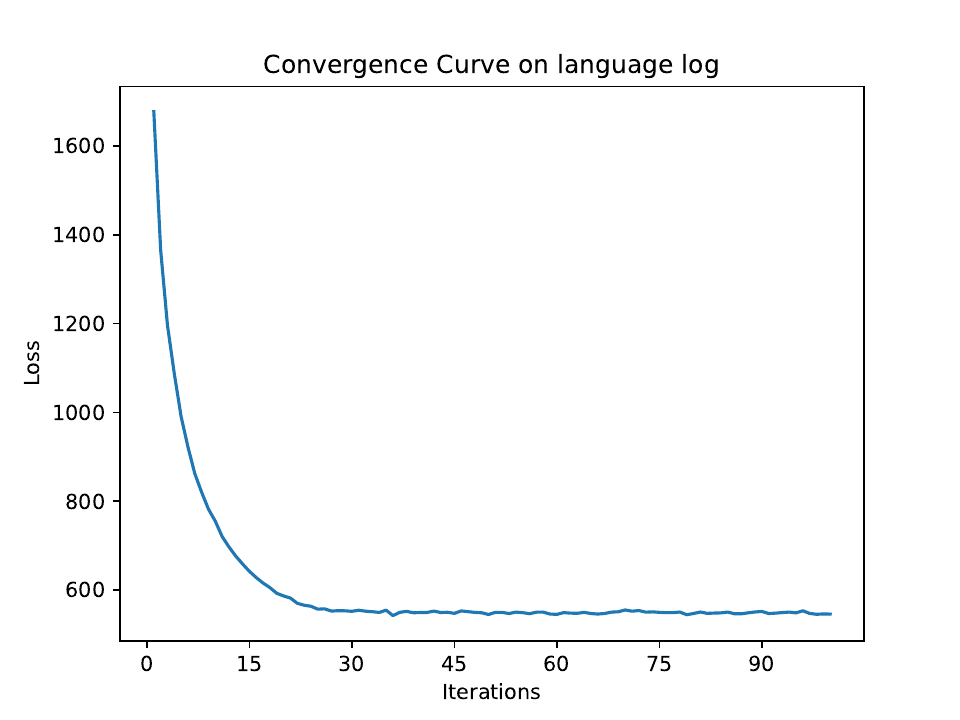}}
	\subfigure[\label{fig:ac}][birds]{
		\includegraphics[scale=0.17]{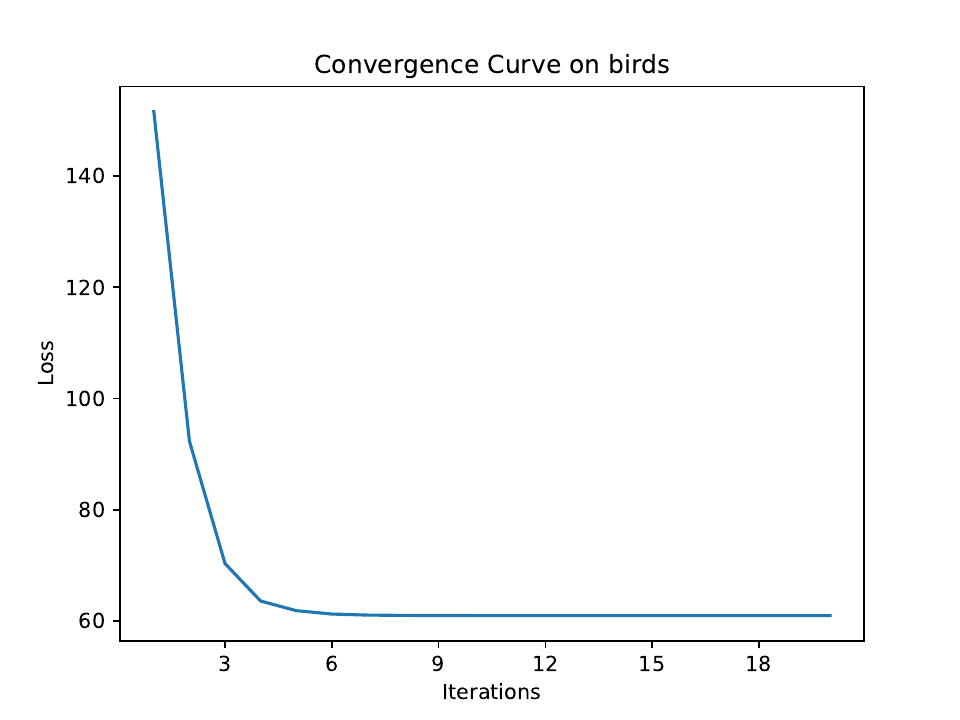}}
	\subfigure[\label{fig:bc}][yeast]{
		\includegraphics[scale=0.17]{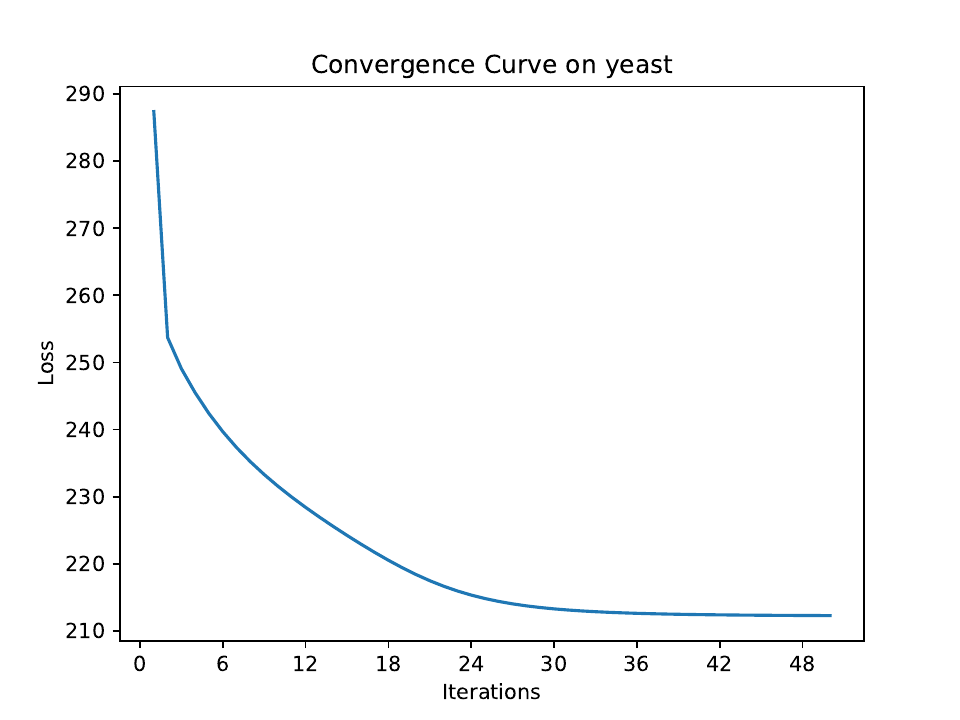}}
	\subfigure[\label{fig:ac}][scene]{
		\includegraphics[scale=0.17]{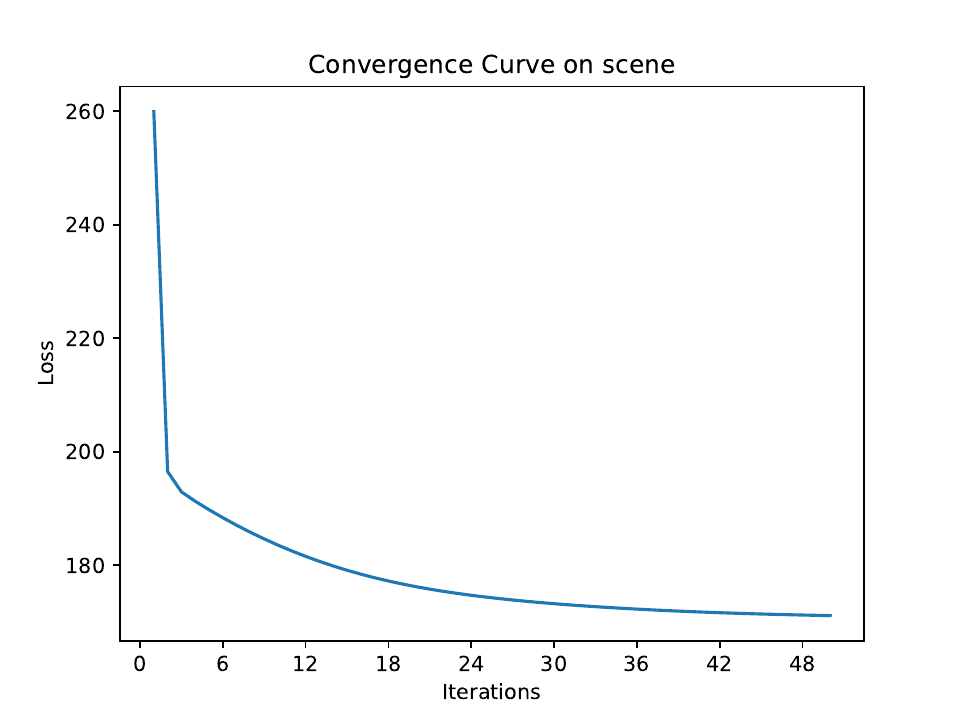}}
	\subfigure[\label{fig:bc}][corel5k]{
		\includegraphics[scale=0.17]{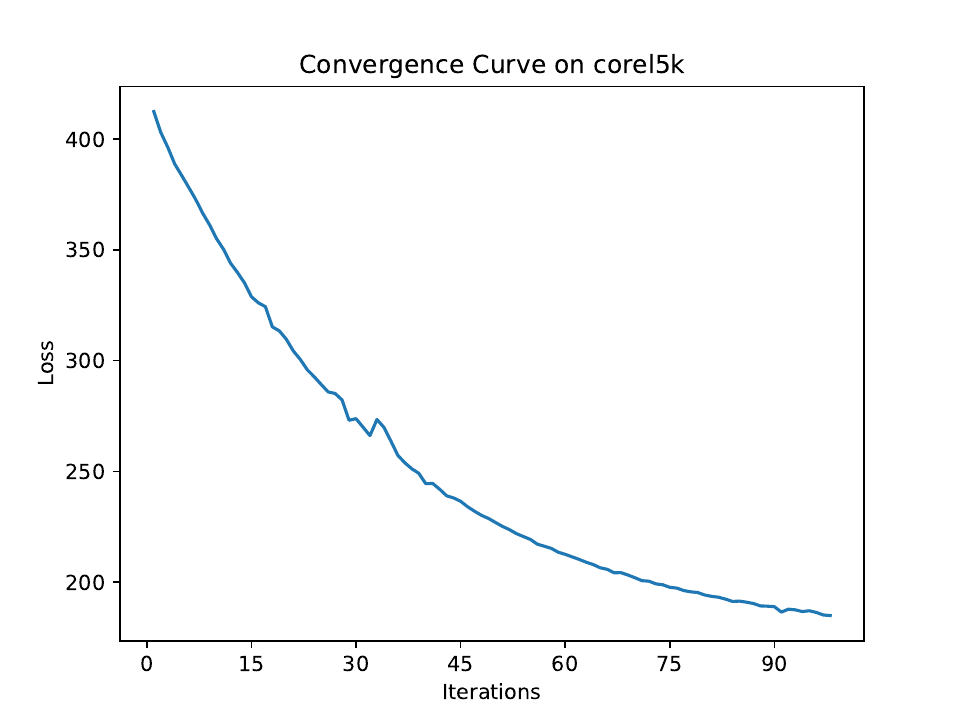}}
	\subfigure[\label{fig:bc}][corel16k]{
		\includegraphics[scale=0.17]{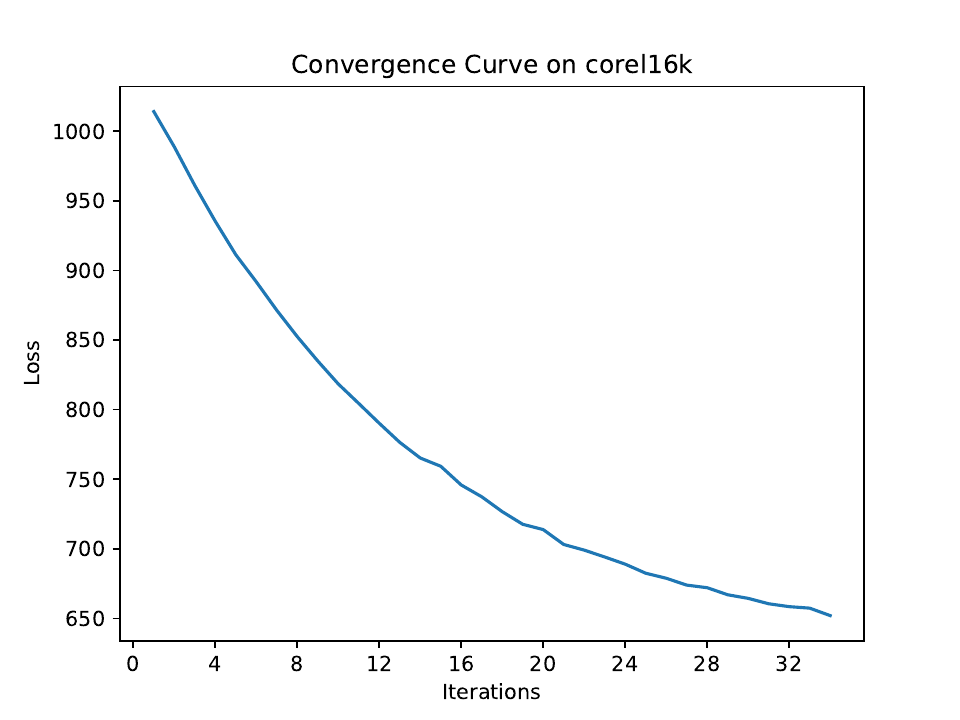}}
		\\
	\caption{Convergence Curve.}
	\label{fig convergence} 
\end{figure*}

\begin{figure*}
    \centering
    \includegraphics[scale=0.5]{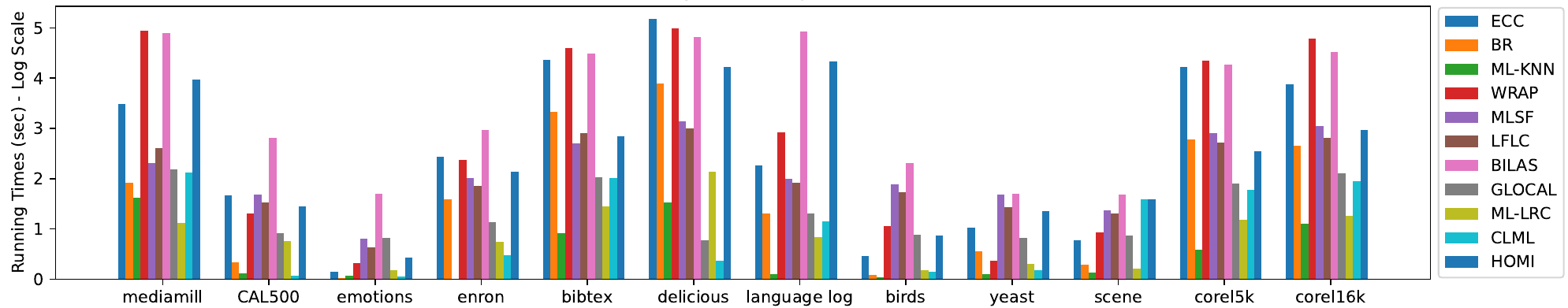}
    \caption{Running time comparison.}
    \label{fig:running time}
\end{figure*}

\begin{table}[ht]
\setlength{\tabcolsep}{3.4mm}\renewcommand\arraystretch{1} 
    \centering
 \caption{Comparison of HOMI against other comparing approaches with \textit{Holm's procedure} w.r.t. all evaluation metrics at significance level $\alpha = 0.05$}
    \begin{tabular}{c c c c c}
     \hline \hline
        \multicolumn{5}{c}{Hamming Loss}\\\hline
        j & approach& $z_j$ & $p_j$ & $\alpha / k - j +1$ \\\hline
        2 & CLML & -4.801 & 1.582e-6 & 0.005 \\
        3 & GLOCAL & -4.308 & 1.645e-4 & 0.006\\
        4 & BR & -3.261 & 1.106e-3 & 0.006\\
        5 & ML-LRC & -3.200 & 1.372e-3 & 0.007\\
        6 & ECC & -3.077 & 2.088e-3 & 0.008\\
        7 & ML-KNN & -2.646 & 8.133e-3 & 0.010\\
        8 & MLSF & -2.646 & 8.133e-3 & 0.013\\
        9 & WRAP & -1.600 & 1.095e-1 & 0.017 \\
        10 & BILAS & -1.538 & 1.238e-1 & 0.025\\
        11 & LFLC & -0.677 & 4.984e-1 & 0.050\\
        \hline

        \multicolumn{5}{c}{Ranking Loss}\\\hline
        j & approach& $z_j$ & $p_j$ & $\alpha / k - j +1$ \\\hline
        2 & ML-LRC & -4.185 & 2.850e-5 & 0.005 \\
        3 & GLOCAL & -3.508 & 4.513e-4 & 0.006\\
        4 & ML-KNN & -2.646 & 8.133e-3 & 0.006\\
        5 & CLML & -1.846 & 6.483e-2 &  0.007\\
        6 & ECC & -1.784 & 7.428e-1 &  0.008\\
        7 & MLSF & -1.354 & 1.757e-1 & 0.010\\
        8 & BILAS & -1.107 & 2.679e-1 & 0.013\\
        9 & LFLC & -0.430 & 6.665e-1 & 0.017 \\
        10 & BR & 0.123 & 1.000 & 0.025\\
        11 & WRAP & 0.492 & 1.000 & 0.050\\
        \hline

     \multicolumn{5}{c}{One-error}\\\hline
        j & approach& $z_j$ & $p_j$ & $\alpha / k - j +1$ \\\hline
        2 & ML-KNN & -3.754 & 1.738e-4 & 0.005 \\
        3 & ECC & -2.277 & 2.277e-2 & 0.006\\
        4 & MLSF & -2.154 & 3.123e-2 & 0.006\\
        5 & GLOCAL & -1.907 & 5.641e-2&  0.007\\
        6 & ML-LRC & -1.723 & 8.483e-1 &  0.008\\
        7 & BILAS & -1.354 & 1.757e-1 & 0.010\\
        8 & CLML & -0.861 & 3.889e-1 & 0.013\\
        9 & BR & -0.800 & 4.237e-1 & 0.017 \\
        10 & LFLC & 1.107 & 1.000 & 0.025\\
        11 & WRAP & 1.538 & 1.000 & 0.050\\
        \hline
        
     \multicolumn{5}{c}{Macro-averaging AUC}\\\hline
        j & approach& $z_j$ & $p_j$ & $\alpha / k - j +1$ \\\hline
        2 & GLOCAL & -5.046  & 4.494e-7 & 0.005 \\
        3 & ECC & -5.046  & 4.494e-7 & 0.006\\
        4 & BR & -4.615 & 3.913e-6 & 0.006\\
        5 & MLSF & -4.554 & 5.254e-6 & 0.007\\
        6 & WRAP & -3.200 & 1.372e-3 & 0.008\\
        7 & BILAS & -3.077 & 2.089e-3 & 0.010\\
        8 & LFLC & -2.831 & 4.638e-3 & 0.013\\
        9 & CLML & -2.154 & 3.123e-2 & 0.017\\
        10 & ML-KNN & -1.354 & 1.757e-1 & 0.025\\
        11 & ML-LRC & -1.292 & 1.962e-1 & 0.050\\
        \hline\hline
    \end{tabular}
   
    \label{tab:holm}
\end{table}

\begin{itemize}
    \item HOMI performs better than all the first-order and second-order methods. For example, HOMI performs significantly better than BR according to Table \ref{tab:holm}. It performs nearly 2 times better than BR on bibtex in average with respect to hamming loss, 1.6 times better than BILAS with respect to macro-averaging AUC on CAL500, and more than 6 times better than ML-KNN on bibtex with respect to one-error, which validates the importance of involving high-order information.
    
    \item HOMI also significantly outperforms all the high-order methods including the low-rank based approaches. Specifically, the improvements of HOMI over GLOCAL and ML-LRC are significant according to Table \ref{tab:holm}, which are two state-of-the-art multi-label classification methods based on low-rank factorization. For example, HOMI performs nearly 2 times better than GLOCAL on corel16k001 regarding macro-averaging AUC and nearly 4 times better w.r.t. ranking loss on bibtex. It outperforms ML-LRC more than 2 times on CAL500 w.r.t. ranking loss. \textit{This observation verifies the rationality of our basic assumption that the label matrix of multi-label classification should be high-rank rather than low-rank.} 
    
    \item LFLC and BILAS perform well on scene, while they are not apt to deal with enron and bibtex. Besides, LFLC performs well on emotions, but does not on mediamill. However, HOMI is adept in almost all kinds of data set especially on text and image data sets, which validates the robustness of HOMI to different types of data sets.
    
    \item HOMI is also robust to different evaluation metrics compared with the other approaches. HOMI achieves especially outstanding performance on macro-averaging AUC, for it enables positive labels to rank higher than negative ones effectively.

    \item The performance of WRAP and LFLC seems comparable to HOME with respect to the Hamming Loss and the One-error, while HOMI is obvious superior to the other methods with respect to the Macro-averaging AUC. The reason is that HOMI uses self-representation to exploit the high-order label correlations while keeping the label matrix full-rank, leading to more effective label-wise discrimination and aggregation. Therefore, HOMI outperforms the other two methods significantly w.r.t. Macro-averaging AUC.
    
    \item In general, HOMI performs superior or at least comparable to the other algorithms in 85\%, 71.7\%, 65.8\%, 90.8\% cases in terms of hamming loss, ranking loss, one-error and macro-averaging AUC which validates that HOMI is a promising approach in multi-label classification.
   
\end{itemize}

The success of HOMI is partially credited to the information of high-order label correlations exploited from label space, and partially credited to the incorporation of the local geometric structure of instances to achieve joint learning of high-order label correlations and model prediction.

\subsection{Further Analysis}

\subsubsection{High-order correlations exploited by HOMI}

HOMI exploits high-order information during the training process. In order to know what information HOMI has learned, we recorded the high-order correlation matrix $\mathbf{B}$ trained on emotions. Emotions is a multi-label data set that describes the emotions of different music. It contains six common emotions, such as happy, relaxing and angry.

Fig. \ref{fig:my_label}(a) shows the matrix $\mathbf{B}$ that HOMI learns. The value of each column represents the contribution of the corresponding row label to the column label. In order to better display the learned high-order label correlations, the matrix is normalized, and the relationships among labels extracted from $\mathbf{B}$ is visually represented in Fig. \ref{fig:my_label}(b). It is obvious that the diagonal element of $\mathbf{B}$ is large, which means in prediction, a label should be dominated by itself. Moreover, it should also be influenced by correlated labels.  We can clearly observe that label ``relaxing/calm'' and ``quiet/still'' are positively related, while ``amazed/surprised'' and ``sad/lonely'' are negatively correlated, and ``happy/pleased'' and ``sad/lonely'' are negatively correlated, too. Additionally, label ``angry/aggressive'' has negative contribution to all the other labels except ``amazed/surprised''. Those label correlations learned by HOMI are consistent with common sense.

We can further verify the correctness and validity of the excavated higher-order information based on the Russell's emotion circumplex theory \cite{russell1980circumplex}. Russell thought that emotions can be measured in two dimensions, i.e., pleasure–displeasure and degree-of-arousal. We show a simplified Russell's emotion circumplex in Fig. \ref{fig:my_label}(c). We assume that the correlations between two emotions can be determined by their cosine similarity, i.e, if the similarity is larger than 0, the two emotions are positively connected and a larger cosine similarity means more correlation; ; otherwise, they are negatively correlated. From Fig. \ref{fig:my_label}(c), it can be observed that ``happy/pleased'' and ``sad/lonely''are negatively related, while ``angry'' and ``relaxing/calm'' are negatively correlated, which is also learned by HOMI.

We can conclude that the correlations exploited by HOMI are reasonable as they are confirmed by both common sense and  the Russell’s emotion circumplex theory. 

\subsubsection{Usefulness of High-order Information}
In order to validate the effectiveness of using high-order information, we compared HOMI with its degenerated version that makes predictions without considering the high-order label correlations, i.e.,  $\mathbf{g}(\mathbf{x})=\mathbf{xW}+\mathbf{1}_n\mathbf{z}^
\mathsf{T}$ on all the twelve data sets with respect to hamming loss, ranking loss, one-error and macro-averaging AUC. The detailed comparisons are shown in Fig. \ref{fig high}.

From Fig. \ref{fig high} we can observe that the performance of HOMI with high-order information outperforms its degenerated version in 83.3\%, with total 48 cases (12 data sets $\times$ 4 evaluation metrics). Especially on the data set bibtex with respect to the one-error, HOMI performs 5 times better than its degenerated version. As the matrix $\mathbf{B}$ captures the high-order correlations among labels, it is effective in improving prediction accuracy. We thus can conclude that considering high-order label information when making predictions enable HOMI to perform better and more stable in general. 

\subsubsection{Effectiveness of Joint Learning}
HOMI exploits the high-order label correlations and make predictions in a joint manner. To show the effectiveness of joint learning, we also designed a compared method with a step-wise learning fashion that first learns high-order information by minimizing $\|\mathbf{Y}-\mathbf{YB}-\mathbf{1}_n\mathbf{t}^\mathsf{T}\|_F^2$ and then makes predictions by $(\mathbf{xW}+\mathbf{1}_n\mathbf{z}^\mathsf{T})\mathbf{B}+ \mathbf{1}_n\mathbf{t}^\mathsf{T} $. 

The comparison between joint learning and step-wise learning is shown in Fig. \ref{fig joint}, where we can observe that the joint learning manner outperforms the step-wise learning significantly. Specifically, the performance of joint learning model is relatively superior to step-wise learning in 85.4\%, with total 48 cases (12 data sets $\times$ 4 evaluation metrics). Especially on the data set mediamill, the Ranking Loss of joint learning manner is 20 times lower than that of the step-wise one.

\subsubsection{Usefulness of Local Geometric Structure}
Aiming to validate the effectiveness of local geometric structure, we evaluated the performance of HOMI without local geometric structure (i.e., $\gamma=0$) in Fig. \ref{fig geometric}, where we can find that the local geometric structure is important to the proposed approach. For example, HOMI performs 13 times better than its simplified version on bibtex with respect to ranking loss. Generally, the one with local geometric structure significantly outperforms the one without that in 95.9\% cases. 

\subsubsection{Sensitivity Analysis}

In Fig. \ref{fig further}, we investigate the sensitivity of HOMI with respect to $\beta$, $\gamma$, $\lambda$ and the number of nearest instances ($s$) on emotions. It is evident that the performance of HOMI is relatively stable and excellent as the value of the parameters change within a reasonable wide range, validating the robustness of HOMI to the hyper-parameters, which serves as a desirable property in practice.

\subsubsection{Convergence Analysis}
Fig. \ref{fig convergence} shows the convergence property of the proposed approach on twelve data sets where we can observe that the objective function decreases significantly and converges in about 50 iterations on nearly all the data sets. 

\subsubsection{Running time}
We also compare the running time of each algorithm on each data set, which is shown in Fig. \ref{fig:running time}. It is obvious that HOMI is usually faster than BILAS, WRAP and ECC. Additionally, the running speed of HOMI is also comparable to other methods, which is also a promising property in practice.

\section{Conclusion}
In this paper, we have presented an effective multi-label classification approach. Different from the traditional low-rank based methods, we argue that the label matrix of multi-label classification is full-rank or approximately full-rank, and thus we propose to keep the rank of label matrix unchanged by self-representation. Moreover, by incorporating the local geometric structure of the input, the proposed method can simultaneously make predictions and exploit high-order label correlations. Extensive experiments validate the effectiveness of the proposed approach in learning high-order label correlations, and in incorporating the local geometric structure. As the proposed model can explicitly indicate the high-order label correlations, we have verified HOMI can learn meaningful label correlations. Besides, HOMI also significantly outperforms the state-of-the-art multi-label classification approaches, serving as a promising solution for multi-label classification. In the future, it is interesting to investigate how to extend HOMI to a non-linear version.

\appendices
\section{The Matrix $\mathbf{B}$ HOMI Has Learned}
\begin{figure}
    \centering

    \includegraphics[scale=0.54]{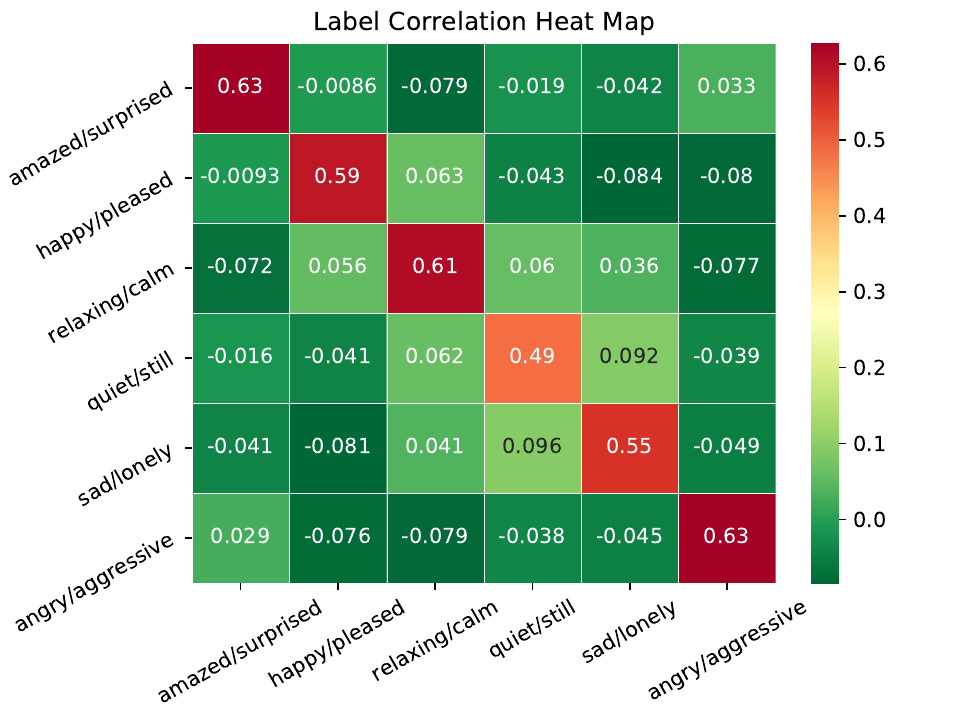}
    \caption{Matrix $\mathbf{B}$ learned by HOMI on the emotion data set. }
    \label{fig:my_label_origin}
\end{figure}

In section 4.6.1, we have recorded the normalized matrix $\mathbf{B}$ trained by HOMI on emotions. The original value of the matrix $\mathbf{B}$ is shown in Fig. \ref{fig:my_label_origin}.

\ifCLASSOPTIONcaptionsoff
  \newpage
\fi

\bibliographystyle{IEEEtran}
\bibliography{tkde}


\vspace{-10mm}

\begin{IEEEbiography}[{\includegraphics[width=1in,height=1.25in,clip,keepaspectratio]{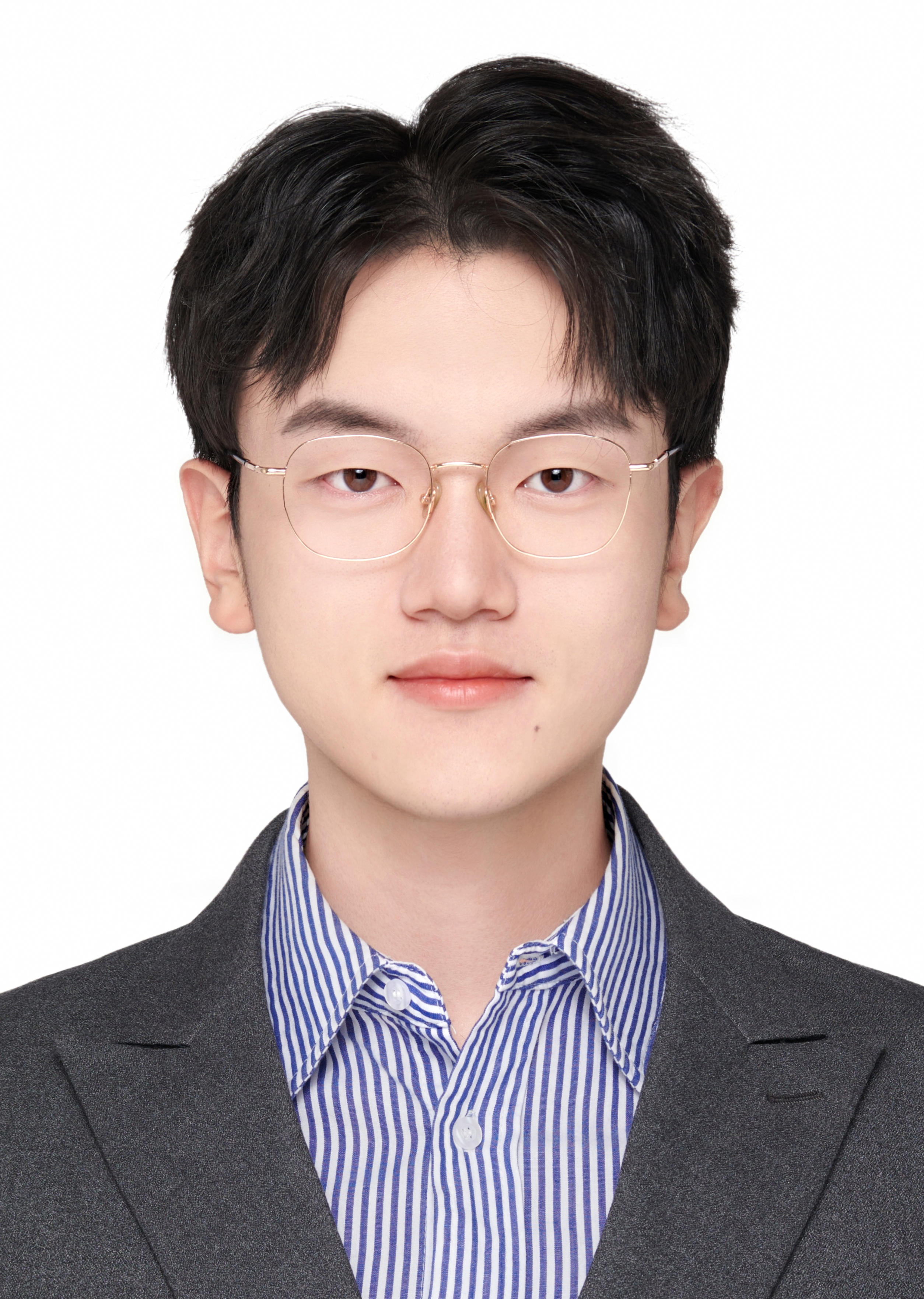}}]{Chongjie Si} received the B.S. degree in artificial intelligence from Chien-Shiung Wu College, Southeast University, Nanjing, China, in 2022. He is currently a Ph.D. student at the Artificial Intelligence Institute, Shanghai Jiao Tong University, Shanghai, China. His current research interests lie in machine learning, incremental learning and semantic segmentation.
\end{IEEEbiography}

\vspace{-10mm}

\begin{IEEEbiography}[{\includegraphics[width=1in,height=1.25in,clip,keepaspectratio]{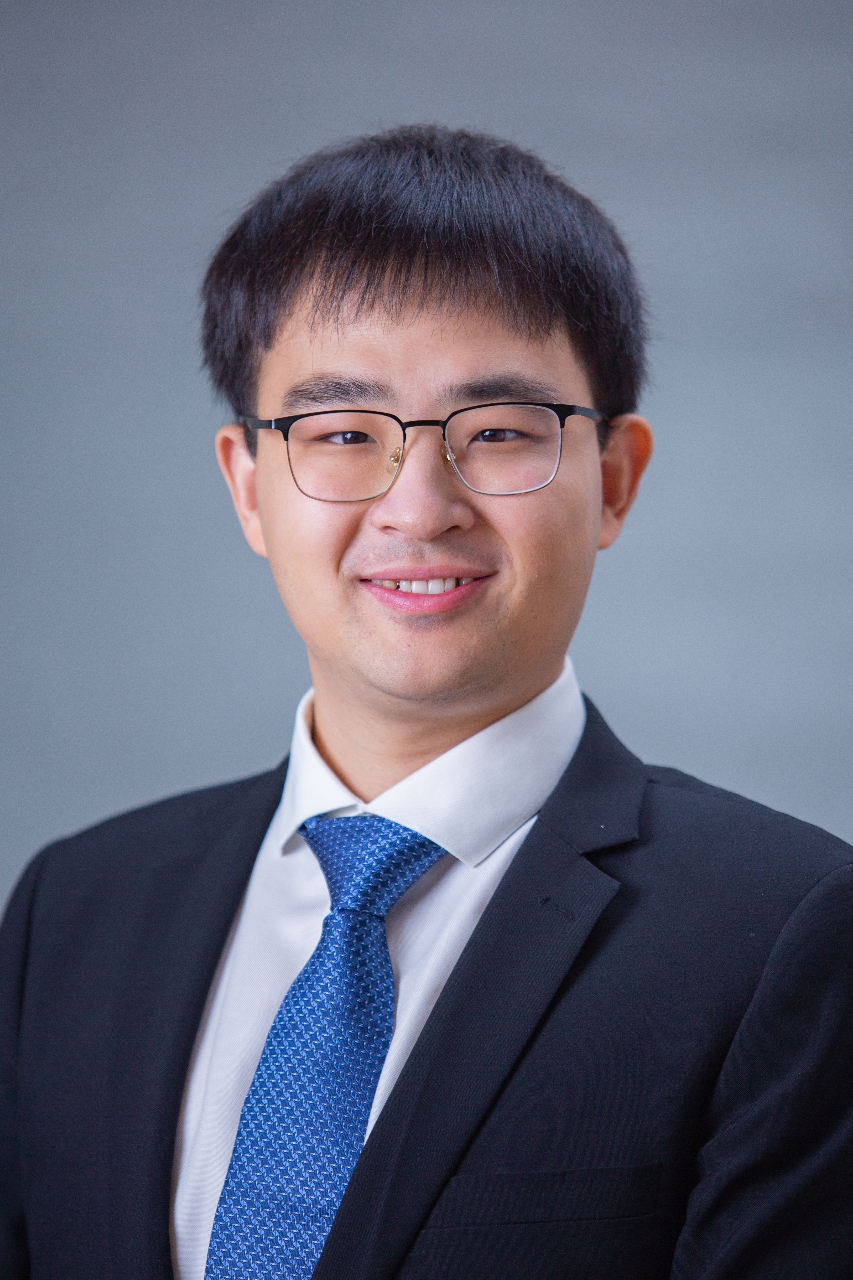}}]{Yuheng Jia} (Member, IEEE) received the B.S. degree in automation and the M.S. degree in control theory and engineering from Zhengzhou University, Zhengzhou, China, in 2012 and 2015, respectively, and the Ph.D. degree in computer science from the City University of Hong Kong, Hong Kong, China, in 2019. He is currently an Associate Professor with the School of Computer Science and Engineering, Southeast University, Nanjing, China. His research interests broadly include topics in machine learning and data representation, such as weakly-supervised learning, high-dimensional data modeling and analysis, and low-rank tensor/matrix approximation and factorization.
\end{IEEEbiography}

\vspace{-10mm}

\begin{IEEEbiography}[{\includegraphics[width=1in,height=1.25in,clip,keepaspectratio]{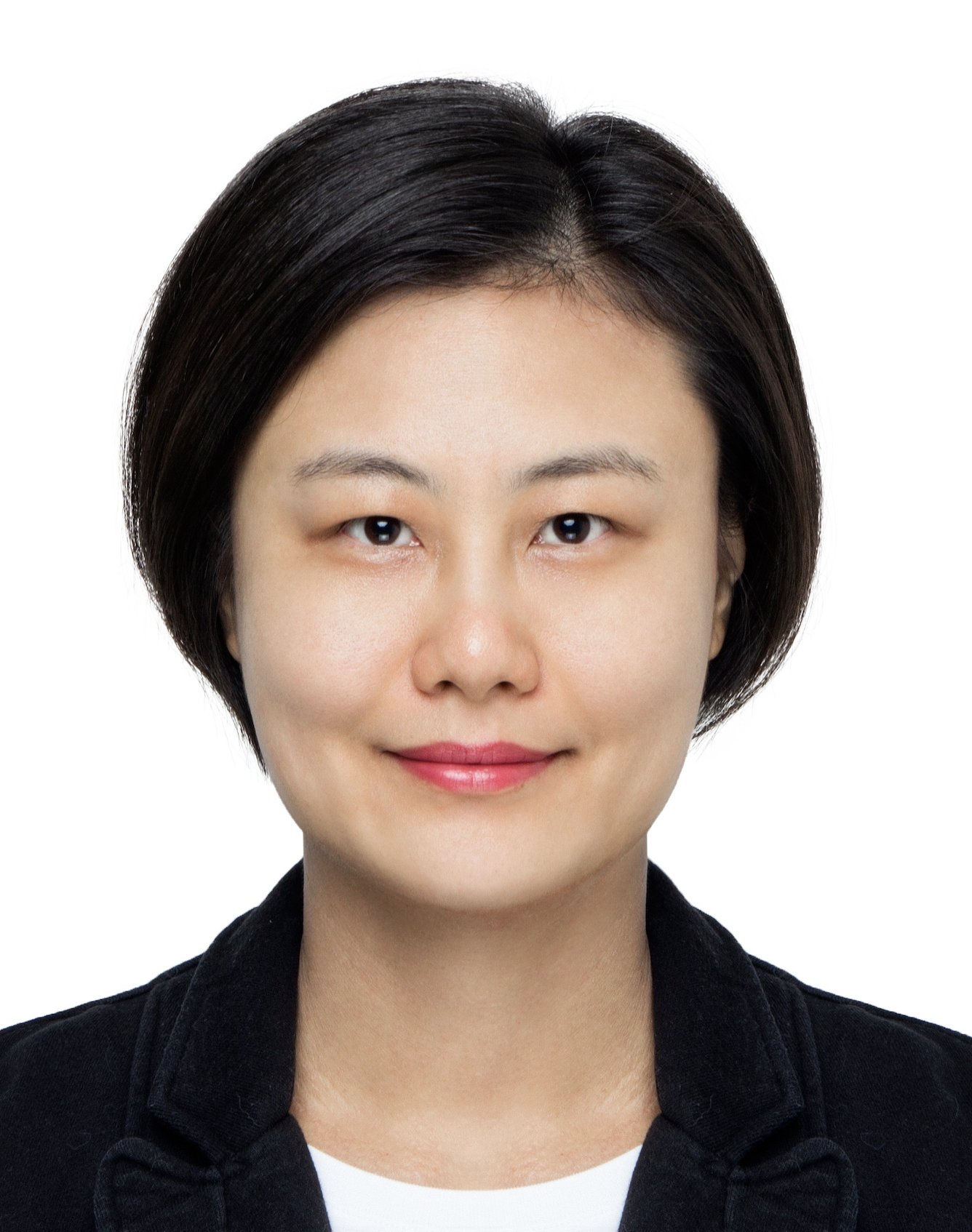}}]{Ran Wang} (Senior Member, IEEE) received her B.Eng. degree in computer science from the College of Information Science and Technology, Beijing Forestry University, Beijing, China, in 2009, and the Ph.D. degree from the Department of Computer Science, City University of Hong Kong, China, in 2014. From 2014 to 2016, she was a Postdoctoral Researcher at the Department of Computer Science, City University of Hong Kong. She is currently an Associate Professor at the School of Mathematical Science, Shenzhen University, China. Her current research interests include machine learning, pattern recognition, uncertainty modeling, fuzzy sets and fuzzy logic, and their related applications.
\end{IEEEbiography}
\vspace{-10mm}

\begin{IEEEbiography}[{\includegraphics[width=1in,height=1.25in,clip,keepaspectratio]{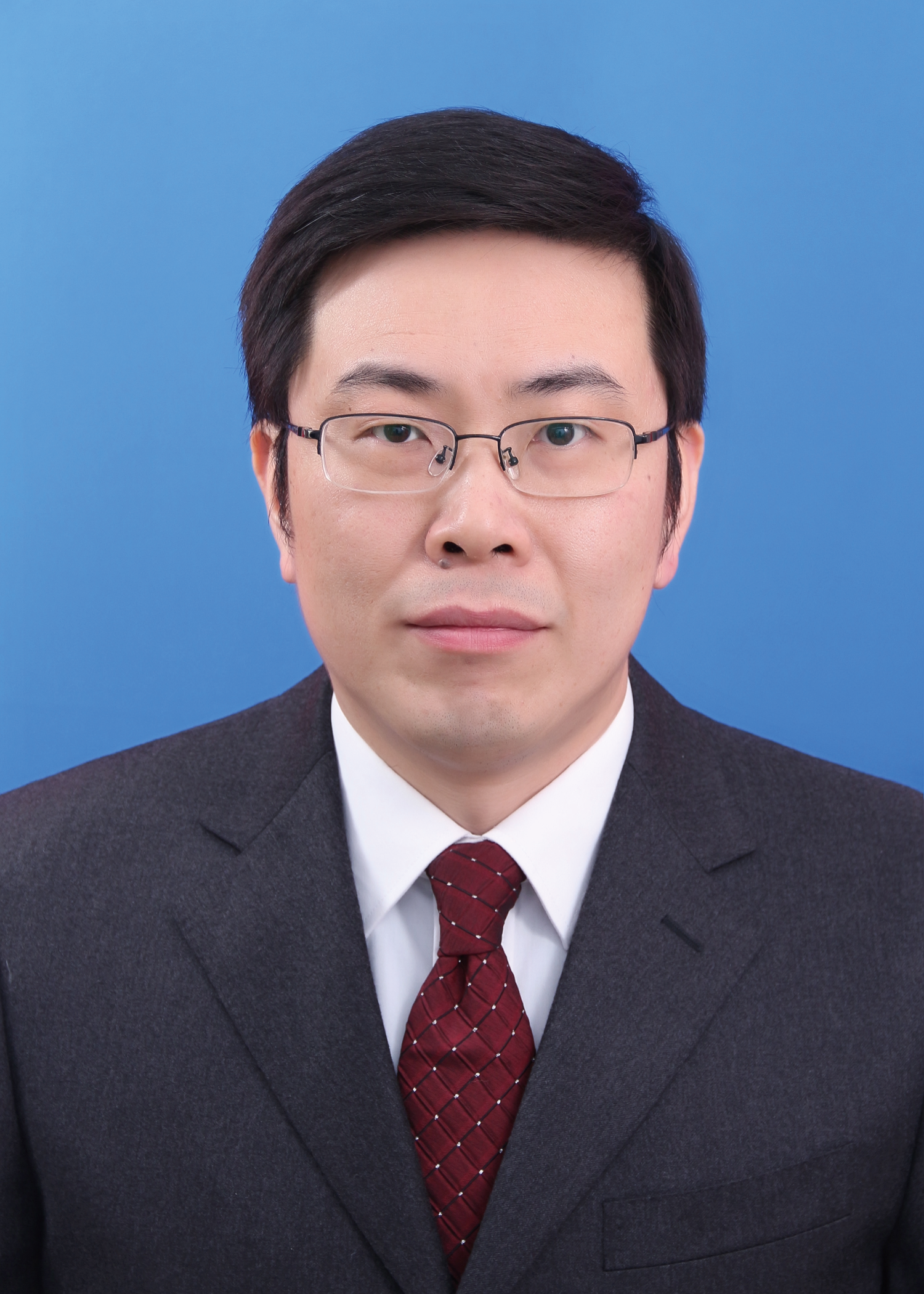}}]{Min-Ling Zhang} (Senior Member, IEEE) received the BSc, MSc, and PhD degrees in computer science from Nanjing University, China, in 2001, 2004 and 2007, respectively. Currently, he is a Professor at the School of Computer Science and Engineering, Southeast University, China. His main research interests include machine learning and data mining. In recent years, Dr. Zhang has served as the General Co-Chairs of ACML’18, Program Co-Chairs of PAKDD’19, CCF-ICAI’19, ACML’17, CCFAI’17, PRICAI’16, Senior PC member or Area Chair of AAAI 2022-2024, IJCAI 2017-2023, KDD 2021-2023, ICDM 2015-2022, etc. He is also on the editorial board of IEEE Transactions on Pattern Analysis and Machine Intelligence, ACM Transactions on Intelligent Systems and Technology, Science China Information Sciences, Frontiers of Computer Science, Machine Intelligence Research, etc. Dr. Zhang is the Steering Committee Member of ACML and PAKDD, Vice Chair of the CAAI Machine Learning Society. He is a Distinguished Member of CCF, CAAI, and Senior Member of AAAI, ACM, IEEE.
\end{IEEEbiography}

\begin{IEEEbiography}[{\includegraphics[width=1in,height=1.25in,clip,keepaspectratio]{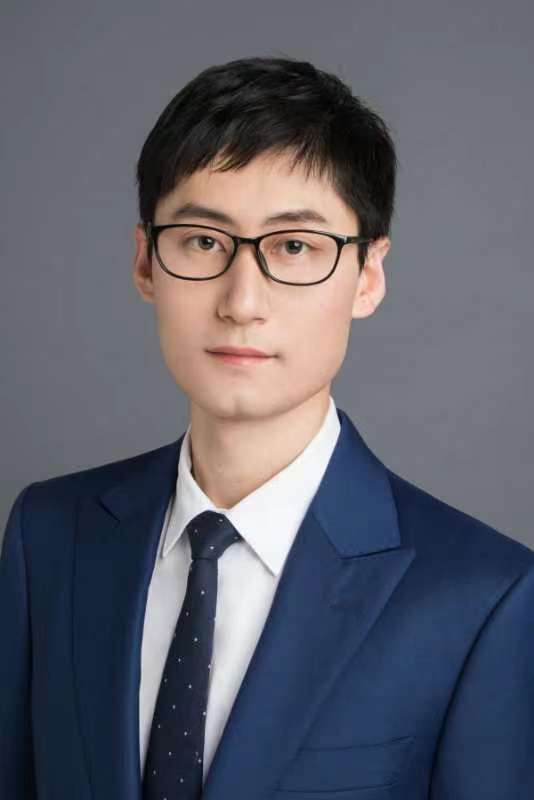}}]{Yanghe Feng} received the master’s and Ph.D. degrees from the Information System and Engineering Laboratory, National University of Defense Technology. He is currently an Associate Professor with the National University of Defense Technology. Before joining the National University of Defense Technology, he was a Visiting Scholar and a Research Assistant Professor with The University of Iowa and Havard University. His Ph.D. research focused on building ``The plan online and learn offline'' framework to enable computers with the ability to analyze, recognize, and predict real-world uncertainty. His primary research interests include casual discovery and inference, active learning, and reinforcement learning. He has published a lot of influential papers in top-tier journals and conferences, e.g., IEEE Transactions on Neural Networks and Learning Systems, IEEE Transactions on Cybernetics, IEEE Transactions on Communications, and so on. He was awarded the young distinguished research scientist of China Command and Control Society. He has been serving as the vice secretary of the Intelligent Computing Subcommittee of the Operational Research Society of China and (co-)editing several international journals.
\end{IEEEbiography}

\begin{IEEEbiographynophoto}{Chongxiao Qu} is with
the 52nd Research Institute of China Electronics Technology Group Corporation, Hangzhou, P.R. China. His research interests lie in deep learning (artificial intelligence), learning (artificial intelligence), autonomous aerial vehicles, computer based training, computer vision, cooperative systems, decision making, image classification, image segmentation, military computing, multi-agent systems, multi-robot systems, etc.
\end{IEEEbiographynophoto}


\end{document}